\documentclass[acmsmall]{acmart}
\AtBeginDocument{%
  }

\setcopyright{rightsretained}
\acmJournal{PACMMOD}
\acmYear{2025} \acmVolume{3} \acmNumber{1 (SIGMOD)}
\acmArticle{71} \acmMonth{2} \acmPrice{}\acmDOI{10.1145/3709721}

\newcommand\modelName{RLER-TTE}
\usepackage{tikz}
\usepackage{enumitem}
\usepackage[ruled,vlined,linesnumbered]{algorithm2e}
\usepackage{algpseudocode}
\usepackage{multirow}
\usepackage{multicol}
\usepackage{stfloats} 
\usepackage{booktabs}
\usepackage{graphicx} 

\newcommand*\circled[1]{\tikz[baseline=(char.base)]{
            \node[shape=circle,draw,inner sep=0.1pt] (char) {#1};}}
\newcommand{\revise}[1]{{#1}}

\usepackage{array}
\usepackage{svg}
\usepackage{subcaption}
\usepackage{tikz}
\usepackage{pgfplots}
\usepackage{verbatim}
\usepackage{amsmath,bm}
\usepackage{tkz-euclide}
\usepackage{marginnote}
\usepackage{tcolorbox}
\tcbuselibrary{breakable}
\usepackage{diagbox}
\usepackage{xcolor}
\usepackage[ruled,linesnumbered]{algorithm2e}

\begin{document}

\title[RLER-TTE: An Efficient and Effective Framework for En Route Travel Time Estimation \\ with Reinforcement Learning]{RLER-TTE: An Efficient and Effective Framework for En Route Travel Time Estimation with Reinforcement Learning}

\author{Zhihan Zheng}
\authornote{Both authors contributed equally to this research.}
\affiliation{%
  \institution{Beijing University of Posts and Telecommunications}
  \city{Beijing}
  \country{China}
}
\email{26510036@bupt.edu.cn}

\author{Haitao Yuan}
\authornotemark[1]
\affiliation{%
  \institution{Nanyang Technological University}
  \city{Singapore}
  \country{Singapore}
}
\email{haitao.yuan@ntu.edu.sg}

\author{Minxiao Chen}
\affiliation{%
  \institution{Beijing University of Posts and Telecommunications}
  \city{Beijing}
  \country{China}
}
\email{chenminxiao@bupt.edu.cn}

\author{Shangguang Wang}
\affiliation{%
  \institution{Beijing University of Posts and Telecommunications}
  \city{Beijing}
  \country{China}
}
\email{sgwang@bupt.edu.cn}

\renewcommand{\shortauthors}{Zhihan Zheng, Haitao Yuan, Minxiao Chen, and Shangguang Wang}


\begin{abstract}
En Route Travel Time Estimation (ER-TTE) aims to learn driving patterns from traveled routes to achieve rapid and accurate real-time predictions. However, existing methods ignore the complexity and dynamism of real-world traffic systems, resulting in significant gaps in efficiency and accuracy in real-time scenarios. Addressing this issue is a critical yet challenging task. This paper proposes a novel framework that redefines the implementation path of ER-TTE to achieve highly efficient and effective predictions. Firstly, we introduce a novel pipeline consisting of a Decision Maker and a Predictor to rectify the inefficient prediction strategies of current methods. The Decision Maker performs efficient real-time decisions to determine whether the high-complexity prediction model in the Predictor needs to be invoked, and the Predictor recalculates the travel time or infers from historical prediction results based on these decisions. Next, to tackle the dynamic and uncertain real-time scenarios, we model the online decision-making problem as a Markov decision process and design an intelligent agent based on reinforcement learning for autonomous decision-making. Moreover, to fully exploit the spatio-temporal correlation between online data and offline data, we meticulously design feature representation and encoding techniques based on the attention mechanism. Finally, to improve the flawed training and evaluation strategies of existing methods, we propose an end-to-end training and evaluation approach, incorporating curriculum learning strategies to manage spatio-temporal data for more advanced training algorithms. Extensive evaluations on three real-world datasets confirm that our method significantly outperforms state-of-the-art solutions in both accuracy and efficiency.
\end{abstract}

\ccsdesc[500]{Information systems~Spatial-temporal systems}


\keywords{travel time estimation; spatial-temporal data mining; trajectory; road networks}

\received{July 2024}
\received[revised]{September 2024}
\received[accepted]{November 2024}

\maketitle

\section{Introduction}

\begin{figure}
  \centering  
  \includegraphics[width=0.8\linewidth]{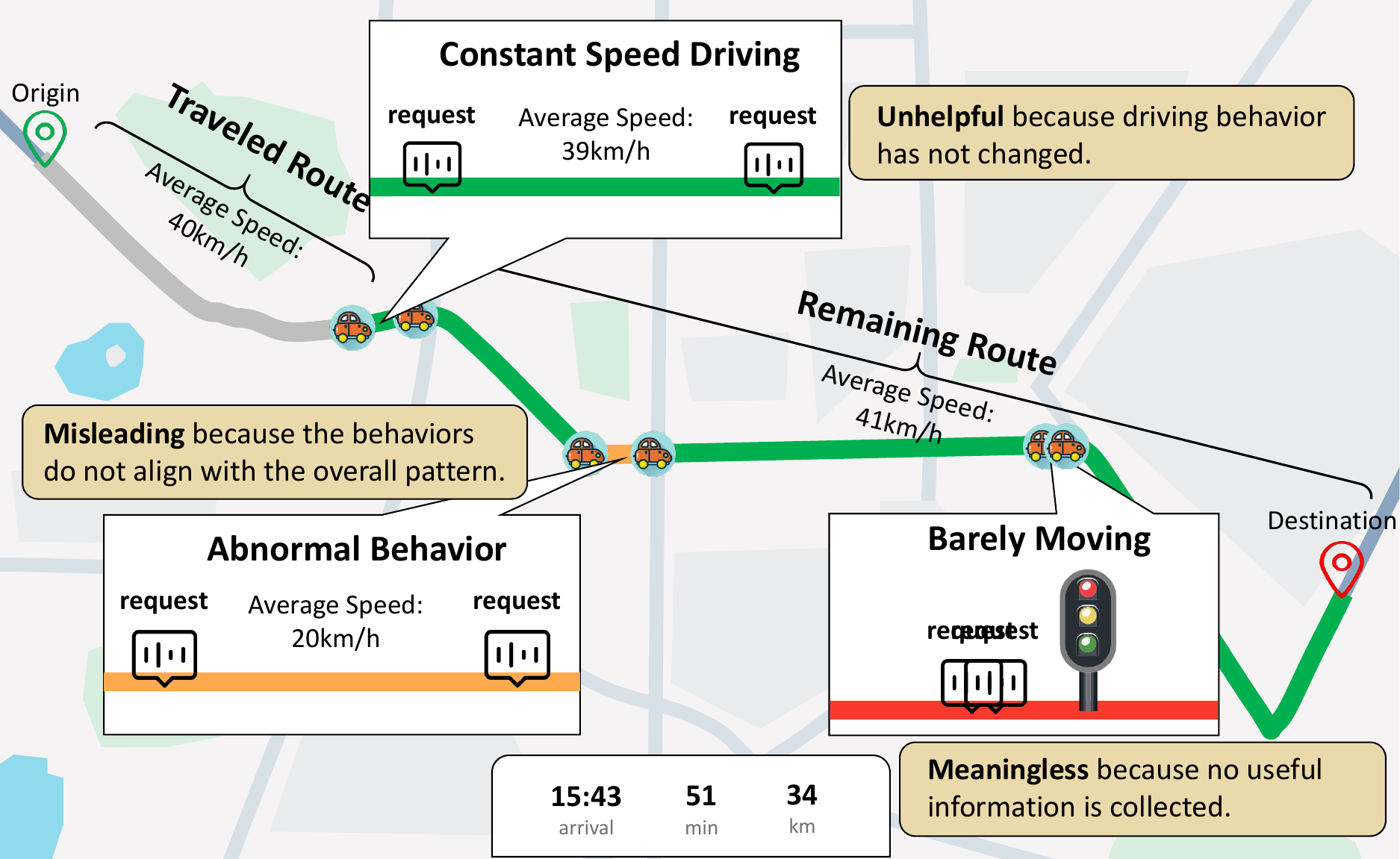}
  \caption{Real-time Requests in En Route Travel Time Estimation when Re-prediction is Ineffective}
  \label{fig:intro}
\end{figure}

Travel Time Estimation (TTE) is one of the most critical operators in intelligent transportation systems, serving as the foundation for tasks such as route planning, vehicle dispatching, and navigation. Traditional TTE methods~\cite{DeepTTE, WDR, ConST, DeepOD, RTTE} provide a one-time estimation of travel time before the journey begins, lacking the capability for meaningful real-time updates during the trip. To address this limitation, the task of En Route Travel Time Estimation (ER-TTE)~\cite{SSML, MetaER-TTE} has been developed, leveraging online data, such as the observed driving behavior of drivers, to enable more accurate online predictions.

Existing ER-TTE methods~\cite{SSML, MetaER-TTE} primarily employ meta-learning techniques to quickly adapt to the unique characteristics of each individual trajectory. In online scenarios, users periodically initiate prediction requests during a journey. These methods receive the requests and utilize deep learning models to recalculate and provide more precise predictions. However, they overlook the complexity and dynamic nature of real-world traffic conditions, leaving critical gaps unaddressed.

\noindent (1) \textbf{Inefficient prediction strategy.} In real-world traffic scenarios, the variability of traffic conditions can lead to the collection of unhelpful and even misleading driving behavior data. As shown in Figure~\ref{fig:intro}, we illustrate three situations where the newly collected data between two consecutive requests is ineffective for travel time estimation. First, when the vehicle is traveling at a constant speed, the newly collected driving behavior remains consistent with the previously learned behavior, offering no additional benefit. Second, there may be a unique road segment that deviates from the general driving pattern before and after, potentially introducing noise and misleading the prediction. Lastly, when the vehicle has barely moved between two requests, no meaningful new data is collected, rendering the data meaningless for further prediction. Additionally, TTE servers in real-world applications handle billions of real-time requests daily. However, existing methods handle these requests by entirely relying on computationally intensive deep learning models for re-calculation. This strategy not only yields minimal improvements in prediction accuracy in many cases but also imposes significant computational stress on the TTE servers. Therefore, it is crucial to develop an innovative solution that can provide accurate real-time predictions while ensuring high efficiency.

\noindent (2) \textbf{Dynamic scenarios and complex data correlations.} To formulate wise prediction strategies, a significant challenge lies in evaluating the quality of online data. Although some simple scenarios are described in Fig.~\ref{fig:intro}, we contend that real-world situations are far more complex. Traffic environments are characterized by dynamic changes and uncertainties, making it difficult to define these problems with explicit rules or models. Moreover, there are intricate spatio-temporal correlations between online data and offline data (e.g., road networks and trajectories). Different road segments exhibit heterogeneous attributes, such as varying speed limits, which correspond to different driving behaviors. However, existing methods for handling these two types of data merely perform simple concatenation, lacking specialized designs to effectively capture the complex correlations between the data. We argue that an ideal solution should thoroughly explore the relationships between the two data types and be capable of making accurate autonomous decisions.

\noindent (3) \textbf{Flawed training and evaluation.} Despite claims of outstanding performance by existing ER-TTE methods, their training and evaluation approaches have certain flaws that fail to reveal their true capabilities in online scenarios. Traditional TTE training methods~\cite{DeepTTE, STANN, ConST} treat an entire trajectory as a single training sample because they only make one prediction at the beginning of the journey. However, in ER-TTE, each trajectory corresponds to numerous prediction requests along the way, with each request having different traveled and remaining routes. Thus, a single trajectory actually corresponds to multiple training samples. Existing methods~\cite{MetaER-TTE, SSML} generate datasets with a fixed proportion of traveled journeys for both training and evaluation. For instance, using a fixed proportion such as 30\% of the traveled route for all trajectories means that each trajectory corresponds to only one prediction request, thus serving as a single training sample. This training strategy results in models that perform poorly when handling prediction requests with different proportions of traveled routes, thereby failing to effectively address all requests for the entire trajectory. Therefore, it is crucial to establish an effective training strategy.

To address the aforementioned challenges, we present a novel framework, \textbf{\modelName}, designed for efficient and robust en route travel time estimation. \textbf{To address the first challenge}, we propose a novel pipeline that redefines the data flow in an online data processing system for ER-TTE. The framework resolves the tradeoff between reusing results and invoking effective models when handling large-scale ER-TTE requests. In online prediction scenarios, the system continuously receives newly collected online data along with prediction requests. Instead of recalculating all requests, the requests first pass through a lightweight \textit{Decision Maker}, which analyzes the driving patterns from the complex online data and determines whether the newly collected information can improve the predictions. Following this decision, a \textit{Predictor} will decide whether to invoke the computationally intensive prediction model or to infer directly from historical prediction results, based on the decision outcomes. Next, \textbf{in response to the second challenge}, we model the online decision-making problem as a Markov Decision Process and carefully design states that effectively represent the complex dynamics of online traffic scenarios, along with a well-defined reward function. We refine the definition of the state to ensure it captures the intricate correlations between online and offline data. Reinforcement learning, specifically Q-learning, with meticulously designed feature encoding techniques and an efficient model architecture, is employed to enable online decision-making. We demonstrate how machine learning techniques can be innovatively applied to online data processing systems, and our strategies can be broadly extended to other similar real-time decision-making systems.
Finally, \textbf{to address the third challenge}, we introduce a curriculum learning strategy and propose an end-to-end training and evaluation method. For each trajectory, we assume that users initiate requests at fixed time intervals, resulting in multiple training samples with varying proportions of traveled segments. This approach generates a large volume of training data. Training in a randomly shuffled order significantly reduces efficiency and slows down convergence. To mitigate this, we design a curriculum by assessing the difficulty of all training samples and selectively arranging the learning sequence. Since the difficulty of samples from trajectories of different lengths cannot be directly compared, we first partition the data into non-overlapping metasets and evaluate them within each metaset. Subsequently, we propose a training scheduling algorithm that arranges the curriculum based on the assessed difficulty.

In summary, the main contributions are listed as follows.
\vspace{-\topsep}
\begin{itemize}[leftmargin=10.2pt]
\setlength{\itemsep}{0pt}
\setlength{\parsep}{0pt}
\setlength{\parskip}{0pt}
\item We introduce a novel framework, \textbf{\modelName}, which for the first time establishes an en route travel time estimation pipeline consisting of two primary components: the \textit{Decision Maker} and the \textit{Predictor}. This framework redefines the data flow of online data processing in the ER-TTE system, significantly enhancing overall system efficiency and performance (Sec.~\ref{sec:3}).
\item We define a Markov decision process to address decision-making challenges in complex real-time scenarios and carefully design the state and reward functions to ensure the effective operation of the system. Additionally, we employ reinforcement learning techniques as a decision-making tool and design feature representation and encoding methods. Our work demonstrates the practical application of ML techniques in solving data management challenges. (Sec.~\ref{sec:4})
\item We propose an end-to-end training method for the Predictor and define a more scientific evaluation methodology, establishing new standards for the ER-TTE system. In addition, we introduce a curriculum learning strategy to manage trajectory data, incorporating two carefully designed components: the Trajectory Difficulty Measurer and the Training Scheduler. These components accelerate model convergence and guide the model towards better minima in the parameter space (Sec.~\ref{sec:5}).
\item We conduct a comprehensive evaluation on three real-world datasets and demonstrate that our approach significantly outperforms existing methods in terms of both accuracy and efficiency. (Sec.~\ref{sec:6})
\end{itemize}

\section{PRELIMINARIES}

We first introduce some basic concepts and then formalize the problem of ER-TTE. 

\subsection{Basic Concepts}
\noindent \textbf{Road Network.} The road network can be formally represented as a directed graph $G=\langle V, E \rangle$, where $V$ denotes the set of vertices corresponding to the endpoints of road segments, and $E$ represents the set of directed edges, each corresponding to a road segment connecting a pair of vertices in $V$.

\noindent \textbf{Travel Route.} A raw trajectory consists of a sequence of GPS points, each denoted as $\langle [\phi_i, \psi_i], t_i \rangle$. Here, \( \phi_i \) and \( \psi_i \) represent the spatial coordinates—latitude and longitude, respectively, and the term \( t_i \) designates the timestamp. We utilize existing map-matching algorithms~\cite{MM1, MM2} that have been broadly adopted in the field~\cite{MMapply,DBLP:journals/pvldb/Wang0L022} to associate these GPS points with specific road segments. Simultaneously, we calculate the travel time for each road segment as $y_i = t_i[-1] - t_i[1]$, where $t_i[1]$ and $t_i[-1]$ denote the start and end timestamp, respectively. Notably, we employ the linear interpolation technique to compute \( t_i \). Consequently, a travel route $\mathcal{R}$ can be represented as a sequence of connected links $\{l_1, l_2, \ldots, l_m\}$, where $l_i = \langle e_i, y_i \rangle$ and $e_i$ is the $i$-th road segment.


\subsection{Problem Definition}

\noindent \textbf{Travel Time Estimation.} We define a Travel Time Estimation (TTE) request as $req = (\mathcal{R}, \rho)$, where $\mathcal{R}$ denotes the travel route and $\rho$ represents the departure time of the route. The objective of TTE is to predict the total travel time for the entire travel route based on the issued request $req$.

\noindent \textbf{En Route Travel Time Estimation.} The ER-TTE task is aimed at real-time prediction in online scenarios. As users drive en route, the ER-TTE system continuously receives prediction requests. 
To clarify, we represent the sequence of requests in chronological order as $[t_1,t_2,...,t_n]$, where $n$ denotes the number of requests. Assuming a request is issued at time $t$, the request can then be formalized as $req = (\mathcal{R}, \rho, t)$, where the route $\mathcal{R}$ can be divided into two parts: the segments already traveled, denoted as $\mathcal{R}_{traveled} = \{l_1, l_2, \ldots, l_{i^{t}-1}\}$, and the segments yet to be traveled, denoted as $\mathcal{R}_{remain} = \{l_{i^{t}}, l_{i^{t}+1}, \ldots, l_m\}$. The actual travel times of the traveled route segments, denoted as $Y_{traveled} = \{y_1, y_2, \ldots, y_{i^{t}-1}\}$, are considered as online information reflecting the user's past driving behavior. Conversely, the time associated with the remaining route, denoted by $Y_{remain} = \{y_{i^{t}}, y_{i^{t}+1}, \ldots, y_m\}$, serves as the output of the ER-TTE task.

\section{Framework Overview}
\label{sec:3}
\begin{figure*}
  \centering  
  \includegraphics[width=\linewidth]{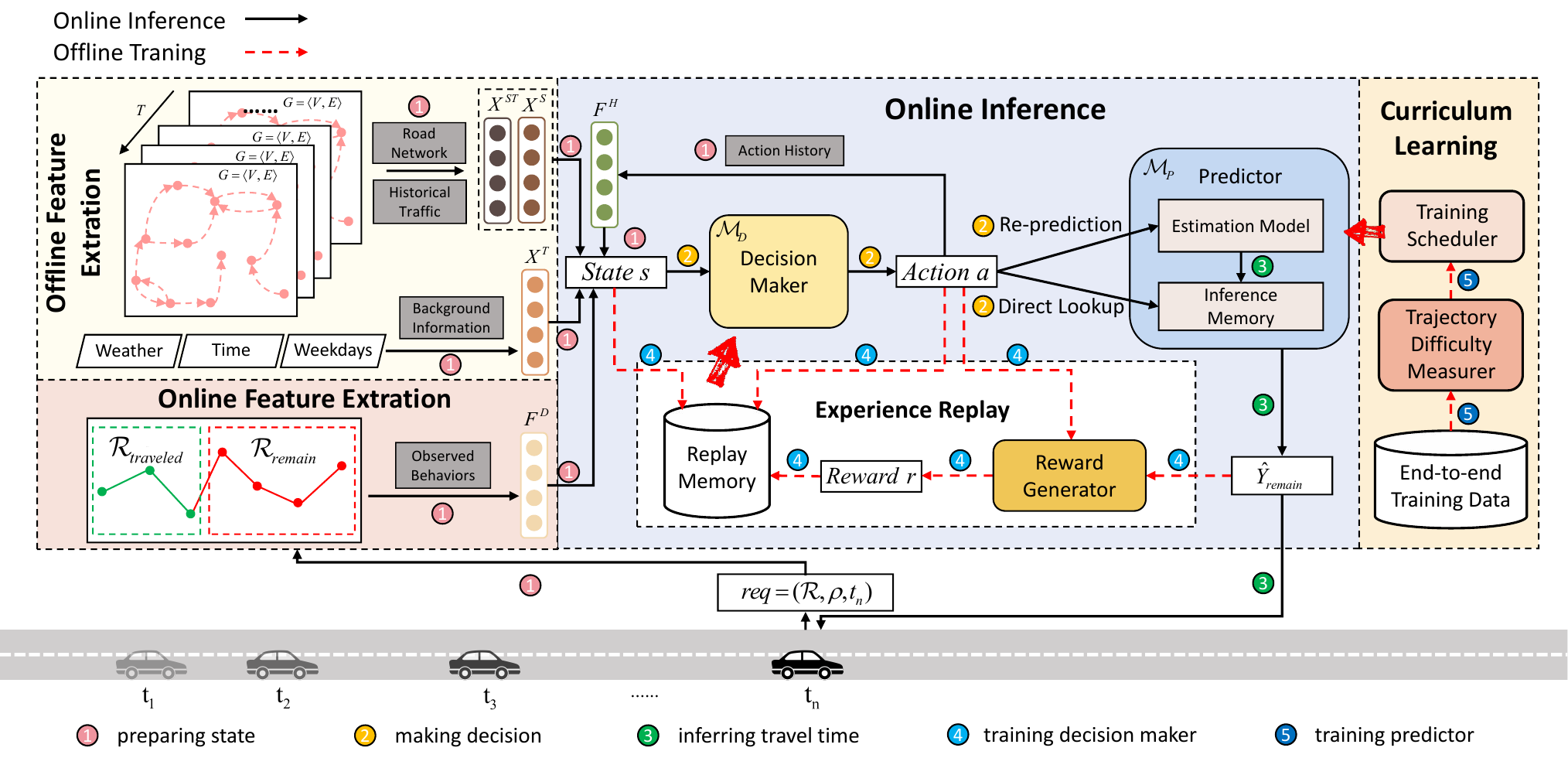}
  \caption{Architecture of the \textbf{\modelName} Framework}
  \label{fig:framework}
\end{figure*}


In this section, we provide a comprehensive description of our system, \textbf{\modelName}. Unlike existing systems that re-predict travel time for each real-time request, our approach incorporates an action selection mechanism to determine whether previous prediction results can be reused. To achieve this, we integrate a reinforcement learning pipeline into the online inference process. As illustrated in Figure~\ref{fig:framework}, our system comprises five modules. The \textit{Offline Feature Extraction} and \textit{Online Feature Extraction} modules are designed to extract relevant features. The \textit{Online Inference} module consists of two deep learning components: \textit{Decision Maker} \(\mathcal{M}_D\) and \textit{Predictor} \(\mathcal{M}_P\). Here, \(\mathcal{M}_D\) decides whether to re-predict new results with \(\mathcal{M}_P\) or use historical prediction results. Finally, \textit{Curriculum Learning} and \textit{Experience Replay} modules are proposed to offline train \(\mathcal{M}_P\) and \(\mathcal{M}_D\). Next, we elaborate on the whole pipeline within our system, detailing the interactions among the five modules.

\smallskip
\noindent \underline{\circled{1} \textbf{Preparing state.}} Suppose at time \( t_n \), the user has a request \( req = (\mathcal{R}, \rho, t_n) \). The \textit{Offline Feature Extraction} module extracts pre-computed spatio-temporal information of the route, which are commonly utilized features in the TTE task~\cite{ConST, STANN, RTTE}. These include spatial-only features (\( X^{S} \)), temporal-only features (\( X^{T} \)), and spatio-temporal features (\( X^{ST} \)). Specifically, the spatial-only features (\( X^{S} \)) are derived from the road network and represent the attributes of road segments (e.g., ID, length, speed limit), as well as their geographical relations. For a given route \(\mathcal{R}\), we define \( X^{S} = \{ x^{s}_{l_i} \}_{l_i \in \mathcal{R}} \), where \( x^{s} \) denotes the spatial attributes of road segment \( l_i \). The temporal-only features (\( X^{T} \)) consist of attributes from the past \( p \) time slots, including departure time, weekdays, and weather conditions. For each segment at the current time \( t_c \), we define \( X^{T} = \{ \{ x^{t} \}^{j}_{l_i} \}^{j \in [t_c - p : t_c]}_{l_i \in \mathcal{R}} \). Similarly, the minimum, maximum, median, and average values of traffic speeds for road segments over the past \( p \) time periods are calculated as historical traffic conditions, which encompass the spatio-temporal features (\( X^{ST} \)). These features are defined as \( X^{ST} = \{ \{ x^{st} \}^{j}_{l_i} \}^{j \in [t_c - p : t_c]}_{l_i \in \mathcal{R}} \), where \( x^{st} = \{ v_{min}, v_{max}, v_{med}, v_{avg} \} \).

Simultaneously, the \textit{Online Feature Extraction} module extracts real-time driving behaviors (\( F^{D} \)) from the traveled route. \( F^D \) are primarily obtained from the actual speed over the traveled road segments, reflecting the user's driving patterns, and is defined as \( F^{D} = \{ f^{d}_{l_i} \}_{l_i \in \mathcal{R}_{traveled}} \). Additionally, we maintain a record, denoted as \( F^{H} \), of the decision history provided by the \textit{Decision Maker}, which is defined as \( F^{H} = \{ f^{h}_{l_i} \}_{l_i \in \mathcal{R}_{traveled}} \). Collectively, these online features and offline features constitute the state \( s \).

\smallskip
\noindent \underline{\circled{2} \textbf{Making decision.}} After feature extraction, the state \( s \), which comprehensively represents the entire route and indicates the current position, is passed to the agent \textit{Decision Maker} \(\mathcal{M}_D\). This agent utilizes a model to generate the action within the reinforcement learning pipeline. Specifically, we employ the Double Deep Q-Network (Double DQN)~\cite{DoubleDQN} as the learning algorithm, which significantly reduces the overestimation of action values and enhances the stability of the learning process in dynamic environments. The network evaluates potential actions and selects the one with the highest value. For actions, we have designed two options: one is to invoke a more complex deep learning model in the \textit{Predictor} for re-prediction, and the other is to directly look up the Inference Memory in the \textit{Predictor}, which continuously stores the most recent prediction results.

\smallskip
\noindent \underline{\circled{3} \textbf{Inferring travel time.}} The \textit{Predictor} \(\mathcal{M}_P\) comprises the Estimation Model and the Inference Memory, which receives and executes the action \( a \). The \textit{Estimation Model}, a sophisticated neural network designed to calculate travel time, employs two state-of-the-art ER-TTE methods, SSML~\cite{SSML} and MetaER-TTE~\cite{MetaER-TTE}, to validate the robustness of our approach. Each time a new calculation is performed, the prediction results are stored in the \textit{Inference Memory}, which can quickly provide prediction results for subsequent requests. The final result, \(\hat{Y}_{remain}\), representing the estimated time for the remaining route, is then conveyed to the user terminal.

\smallskip
\noindent \underline{\circled{4} \textbf{Training decision maker.}} During the training phase, the models within the \textit{Decision Maker} and \textit{Predictor} modules are trained independently. For the \textit{Decision Maker} module, we utilize the widely adopted \textit{Experience Replay} mechanism in reinforcement learning, where an experience is represented as a tuple consisting of state, action, reward, and next state. The processes for generating the state and action have been previously detailed. The reward is derived from the \textit{Reward Generator}, which considers the estimated results $\hat{Y}_{remain}$ and the decision action $a$, comprehensively evaluating efficiency and performance to compute the reward associated with the action. This tuple is then stored in the Replay Memory, which maintains a fixed-size buffer.

\smallskip
\noindent \underline{\circled{5} \textbf{Training predictor.}} In our end-to-end model training method, each trajectory encompasses numerous request instances, resulting in a substantial increase in the quantity of training data and considerable variation in quality. To address this, we introduce a curriculum learning strategy to enhance the training of TTE models within the \textit{Predictor}. Specifically, we employ a \textit{Trajectory Difficulty Measurer} to assess the complexity of predicting a trajectory based on multiple metrics. Subsequently, the \textit{Training Scheduler} devises a curriculum for the \textit{Predictor} to follow.

\section{Online Decision-Making}
\label{sec:4}

In this section, we provide a detailed description of our online decision-making mechanism based on reinforcement learning. Firstly, we discuss how to formulate this problem as a Markov Decision Process (MDP) in Sec~\ref{sec:4.1}. Secondly, we explain the model architecture in Sec~\ref{sec:4.2}, which is specifically engineered to facilitate efficient decision-making within online environments. Lastly, in Sec~\ref{sec:4.3}, we elaborate on the training algorithm employed to optimize the model's performance, accompanied by a complexity analysis.

\subsection{MDP Formulation}
\label{sec:4.1}
\noindent \textbf{State.} At step \( t \), the corresponding state \( s_t \) comprises two types of features: offline features and online features. Offline features (\(X^S, X^T, X^{ST}\)), which can be obtained prior to the start of the journey, reflect the static characteristics of the route. In contrast, online features (\(F^D, F^H\)) are acquired in real-time during the journey. These are utilized to assess the value, neutrality, or potential misleading nature of newly collected online information from the last prediction made by the \textit{Predictor} to the current prediction request. For the \textit{Decision Maker}, online features are crucial for making accurate decisions. Specifically, driving behavior (\(F^D\)) and decision history (\(F^H\)) are pivotal because they collectively capture the driver's unique driving pattern. By incorporating offline features, the \textit{Decision Maker} can further analyze the relationship between these patterns and static route characteristics, thereby enabling more informed decision-making. In contrast, the \textit{Predictor} focuses on leveraging offline features to make accurate travel time predictions. Particularly, the segment length from \( X^S \) and the traffic conditions provided by \( X^{ST} \) form the foundational basis of prediction, while other features further refine the prediction details.

\noindent \textbf{Action.} We define two types of actions: (1) recalculate the travel time for the current request using the \textit{Estimation Model} in \textit{Predictor}; (2) directly provide the estimated value from the \textit{Inference Memory}. For the sake of clarity, these two actions are referred to as ``re-prediction'' and ``direct lookup'' hereafter.

\noindent \textbf{Transition.} In the ER-TTE process, given a state $s_t$ and a corresponding action $a_t$, the transition will move to the state $s_{t+1}$, which incorporates the newly acquired information of the traveled segment along the route. In particular, the online features $F^D$ and $F^H$ would be updated within the new state.

\noindent \textbf{Reward.} The design of the reward function is crucial in reinforcement learning as it significantly impacts the method's effectiveness. To achieve outstanding performance and efficiency, our reward design must comprehensively consider both aspects. Therefore, our reward function is composed of three components: \textbf{performance}, \textbf{efficiency}, and \textbf{frequency}.

The \textbf{performance} reward directly reflects the accuracy of the travel time prediction results. A straightforward idea is to use the prediction result itself as the reward. However, the range of prediction values can vary significantly across different predictions, making it difficult for \textit{Decision Maker} to evaluate the quality of actions effectively. To address this issue, a reward function capable of eliminating the base value of the predictions and amplifying the differential can be introduced to ensure that the reward is consistent and comparable across different predictions. To formalize this, we need to calculate \textit{Predictor}'s prediction value as the re-prediction result, denoted as \(\hat{y}_{rp}\). Simultaneously, we define \(\hat{y}_{dl}\) as the direct lookup result. The difference between these two predictions is used as the reward, which is defined as follows:
\begin{equation}
   r_{p} = \begin{cases} 
    -(\hat{y}_{dl} - \hat{y}_{rp}) &  a_i = 0 \\
    -(\hat{y}_{rp} - \hat{y}_{dl}) &  a_i = 1
\end{cases}
\end{equation}
where $a_i = 0$ denotes direct lookup and $a_i = 1$ denotes re-prediction.

The \textbf{efficiency} reward imposes a certain penalty each time the \textit{Predictor} is used to encourage efficient usage. By setting a penalty factor $\omega_p$, it ensures that when the improvement achieved through re-prediction is limited, the framework can choose to directly provide the prediction result instead. By adjusting the value of the penalty factor, we can trade-off between performance and efficiency. The \textit{efficiency} reward is defined as follows:
\begin{equation}
   r_{e} = \begin{cases} 
    0 &  a_i = 0 \\
    \omega_p &  a_i = 1
\end{cases}
\end{equation}

The \textbf{frequency} reward limits the usage frequency of the \textit{Predictor} to a certain extent, balancing efficiency and performance. This reward mechanism is designed to influence the \textit{Decision Maker}'s behavior, encouraging it to make direct lookup soon after utilizing the \textit{Predictor} and to favor re-prediction after a long period without using the \textit{Predictor}. To achieve this, the reward is formulated based on the elapsed time between the last prediction made by the Predictor and the current request. Our goal is to ensure that the framework remains dataset-agnostic and is not tied to a specific \textit{Predictor}. Therefore, we implemented a linear function to model this reward, which ensures robustness due to its simplicity and stable growth pattern.
\begin{equation}
    r_{f}(\sigma) = \begin{cases}
     0 &  a_i = 0 \\
     \alpha \cdot \sigma + \beta &  a_i = 1
\end{cases}
\end{equation}
Where \(\alpha\) and \(\beta\) are hyper-parameters, \(\sigma\) represents the time interval between the last prediction and the current time. The value of the linear function transitions from negative to positive as \(\sigma\) increases.

Finally, we can evaluate the overall reward as follows:
\begin{equation}
    r = r_{p} + r_{e} + r_{f}(\sigma)
\end{equation}

\subsection{Model Architecture}
\label{sec:4.2}

\begin{figure}
  \centering  
  \includegraphics[width=0.82\linewidth]{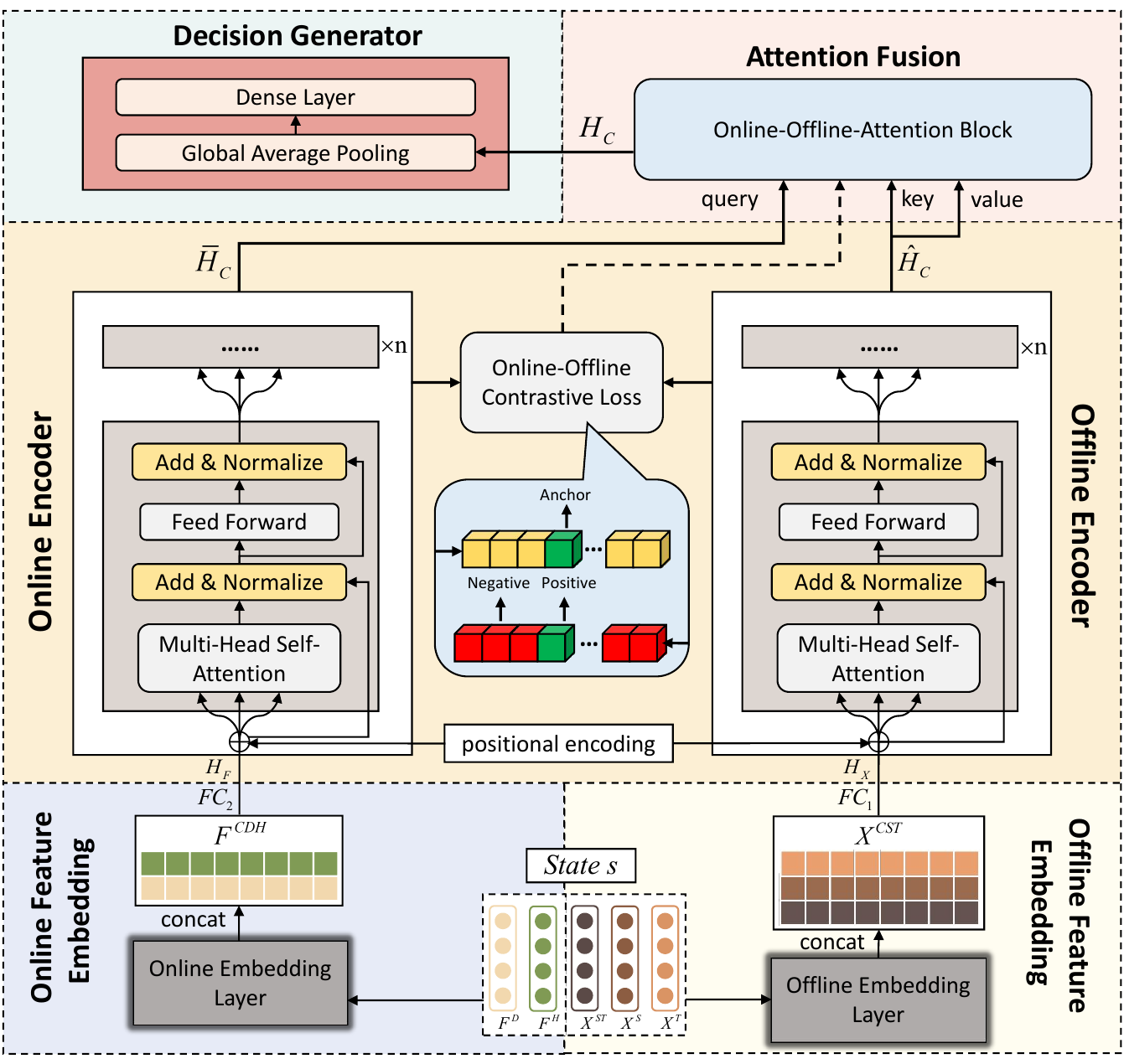}
  \vspace{-0.05in}
  \caption{The Architecture of Decision Maker}
  \vspace{-0.05in}
  \label{fig:dqn}
\end{figure}

In Section~\ref{sec:3}, we provide an overview of the \textit{Decision Maker}, designed to generate decisions efficiently from both online and offline features. As shown in Fig.~\ref{fig:dqn}, we will elaborate on its components, offering a more granular understanding of their structures and functions.

\noindent \textbf{Feature Embedding.} Each road segment is characterized by a series of attributes, including its id, length, and other relevant features. The embedding layer is designed to encode these discrete attribute values into dense vectors. We employ established and widely-used graph embedding techniques~\cite{node2vec} to model the road network's graph structure. The embedding vector for a road segment is denoted as \( h_s \in \mathbb{R}^{d_s} \). Assuming the size of the time slot is defined as \(\Delta T\) minutes, we segment them into weekly cycles, obtaining their embeddings \( h_t \in \mathbb{R}^{\frac{7 \times 24 \times 60}{\Delta T}} \). For the traffic conditions, we first embed the data according to the time slots. Subsequently, we incorporate the historical time slots to obtain the final embedding representation, denoted as \( h_{st} \in \mathbb{R}^{T \times d_{st}} \), where \( T \) is the number of historical time slots. Finally, we concatenate these embeddings and pass them through a fully connected layer ($FC_1$) with ReLU as the activation function to obtain the dense vector of offline features: $H_X = Dense(concat(h_s, h_t, h_{st}))$.

The embeddings of driving behaviors and decision history are denoted by \(h_d\) and \(h_h\). Through another fully connected layer (\(FC_2\)), we embed them into \(H_F\), which is formulated as $H_F = Dense(concat(h_d, h_h))$.

\noindent \textbf{Offline Encoder \& Online Encoder.} To efficiently capture the internal correlations of both online and offline features, we design two modules to encode them separately. These two modules share a similar structure of attention blocks, as the attention mechanism is more efficient in handling sequential data and has demonstrated exceptional performance in capturing the spatio-temporal correlations~\cite{STAN, STAttention}. For the sake of clarity, we will illustrate the architecture using the offline feature encoder as an example. Specifically, each block comprises the following components:

\textit{Multi-Head Self-Attention:} The encoder leverages the Multi-Head Self-Attention mechanism to capture intricate dependencies within the input features \( H_X \). Firstly, we define the query (\( Q_c = \mathbf{W}_q H_X \)), the key (\( K_c = \mathbf{W}_k H_X \)), and the value (\( V_c = \mathbf{W}_v H_X \)), where \( \mathbf{W}_q \), \( \mathbf{W}_k \), and \( \mathbf{W}_v \) are learnable parameters. The self-attention score is computed by \( score(H_X, H_X) = softmax\left(\frac{Q_c \cdot (K_c)^\top}{\sqrt{d_m}}\right) \), where \( d_m \) denotes the size of vector \( Q_c \). Subsequently, we obtain the new encoded representation \( Z_c = score(H_X, H_X) \cdot V_c \). This mechanism allows the model to attend to different parts of the input sequence simultaneously, thereby enhancing the representation capability.

\textit{Add \& Normalize:} This layer comprises two key operations: residual connection and layer normalization. The residual connection is employed to mitigate the vanishing gradient problem, while layer normalization standardizes the inputs to have zero mean and unit variance. The process can be represented as $H_N = LN(H_X \oplus Z_c)$.

\textit{Feed Forward:} To further transforming the encoded representations, we design a fully connected layer, which can be expressed as $Z_f = ReLU(\mathbf{W}_f H_N + b_f)$, where $\mathbf{W}_f$, and $b_f$ are learnable parameters and $ReLU(\cdot)$ is the activation function.

\textit{Add \& Normalize:} Similar to the previous layers, we further enhance the encoding: $\hat{H}_c = LN(H_N \oplus Z_f)$.

The aforementioned four steps constitute the fundamental block of the encoder. By employing these encoder blocks, we obtain representations for both online and offline features, denoted as \( \overline{H}_c \) and \( \hat{H}_c \), respectively.

\noindent \textbf{Contrastive Learning.} Contrastive learning aims to align online features and offline features, thereby enhancing the encoders’ capacity to learn superior feature representations. For a batch of training examples, feature pairs of the same sample are considered positive pairs, while pairs from different samples are negative pairs. We first project the online feature \(\overline{H}_C\) and the offline feature \(\hat{H}_C\) into a higher-dimensional space through linear transformations to obtain the normalized representations, denoted as \(g_n(\overline{H}_C)\) and \(g_f(\hat{H}_C)\). To quantify the similarity between the online and offline features, we define a similarity function $S_{ij} = g_n(\overline{H}_C^{(i)})^\top g_f(\hat{H}_C^{(j)})$, where $i$ and $j$ index the samples in the batch. The alignment between the online and offline features is reinforced using the InfoNCE loss, which aims to maximize the similarity scores between positive pairs. The InfoNCE loss \(L_c\) is computed as follows:
\begin{equation}
    \mathcal{L}_c = -\frac{1}{N} \sum_{i=1}^{N} \log \frac{\exp(S_{ii} / \tau)}{\sum_{j=1}^{N} \exp(S_{ij} / \tau)}
\end{equation}
where \(N\) denotes the batch size, and \(\tau\) represents the temperature parameter. 

\noindent \textbf{Online-Offline-Attention Fusion.} To obtain a comprehensive representation of the current state, we apply an attention mechanism to further integrate the online and offline features. Specifically, the online feature \( \overline{H}_c \) serves as the query, while the offline feature \( \hat{H}_c \) is used as both the key and value. The fusion process is summarized by the following formula: $H_c = softmax\left(\frac{\overline{H}_c \cdot \hat{H}_c^\top}{\sqrt{d_a}}\right) \hat{H}_c$, where \(d_a\) denotes the dimension of the feature space. Thus, the online feature \(\overline{H}_C\) guides the integration of the offline feature \(\hat{H}_C\), allowing the model to focus on the most relevant aspects of the offline information, effectively combining the feature representations into a coherent form.

\noindent \textbf{Decision Generator.} In the final step, we utilize Global Average Pooling followed by a Dense Layer to aggregate the feature representations and predict action values: $Q = Dense(Avg(h_c))$, where $Q$ represents the value associated with each possible action, and the action with the highest value is selected as the output.

\begin{figure}[!t]
\begin{algorithm}[H]
    \label{alg:ml}
    \small 
    \SetKwInOut{Input}{Input}
    \SetKwInOut{Output}{Output}
    \Input{training inputs $\mathbf{Z}$, training labels $\mathbf{Y}$, the Decision Maker $\mathcal{M}_{dc}$, the Predictor $\mathcal{M}_{p}$, experience buffer $\mathcal{B}$, learning rate $lr$, training epochs $ep$, batch size $bs$, loss weight $\lambda$, discount factor $\gamma$, update step $us$, train step $ts$.}
    \Output{parameters $\theta_{dc}$ for the Decision Maker $\mathcal{M}_{dc}$.}
    \caption{Model Learning for \textbf{\modelName}}
    
    $X, F \gets$ generating online and offline features\;
    initialize $\theta_{dc}$ with normal distribution\;
    \For{$m \gets 1 \ldots ep$}{
        $\theta_{dc} \gets ModelTrain(\mathbf{Z}, \mathbf{Y}, \mathcal{B}, X, F, lr, bs, \lambda, \gamma, ts)$\;
        update the main network with $\theta_{dc}$\;
            \If{$m\mod us = 0$}{
                update the target network\;
            }
    }
\end{algorithm}
\begin{algorithm}[H]
    \small 
    \SetAlgorithmName{Function}{function}{List of Functions}
    \renewcommand{\thealgocf}{}  
    \SetKwInOut{Input}{Input}
    \Input{$\mathbf{Z}, \mathbf{Y}, \mathcal{B}, X, F, lr, bs, \lambda, \gamma, ts$}
    \caption{ModelTrain}
    iterations $I = |\mathbf{Z}|$\;
    
    \For{$i \gets 1 \ldots I$}{
        $s_i \gets X_i, F_i$\;
        $a_i \gets (\mathcal{M}_{emb}, \mathcal{M}_{enc}, \mathcal{M}_{atf}, \mathcal{M}_{dcg})(s_i)$\;
        $\hat{Y}_i \gets \mathcal{M}_{p}(X_i^{S}, X_i^{T}, X_i^{ST}, F_i^{D})$\;
        $r_i \gets$ using Equations 1-4 with $a_i$ and $\hat{Y}$\;
        $s_{i+1} \gets X_{i+1}, F_{i+1}$\;
        construct transition $T_i = \{s_i, a_i, r_i, s_{i+1}\}$\ and store into $\mathcal{B}$\;
        \If{$i\mod ts = 0$}{
            random sample transitions $T$ from $\mathcal{B}$ based on $bs$\;
            training iterations $TI = |T|$\;
            \For{$j \gets 1 \ldots TI$}{
                compute $q, q_{target}$ with main and target network\;
                $\mathcal{L} \gets$ using Equations 5, 6, 7 and $\lambda$\;
            }
        }
    }
    return $\theta_{dc}$;
\end{algorithm}
\vspace{-0.1in}
\end{figure}

\subsection{Model Learning}
\label{sec:4.3}

As shown in Algorithm~\ref{alg:ml}, we delineate the process of model learning  for \textbf{\modelName}. We first extract the offline features \(X\) and online features \(F\). We then initialize the parameters of all modules using a normal distribution.We train the model for the specified number of epochs \(ep\). During each epoch, we execute the \(\textbf{ModelTrain}\) procedure to optimize the main network and periodically update the target network according to the update step \(us\).

\noindent \textbf{Model Train.} In each iteration, we utilize \(X_i\) and \(F_i\) to construct the state \(s_i\). Subsequently, the state \(s_i\) is processed by the model architecture described in Section 4.2 to produce the action \(a_i\). Concurrently, the Predictor module produces the new prediction \(\hat{Y}_i\). The reward \(r_i\) is computed based on \(a_i\) and \(\hat{Y}_i\), using Equations 1-4. Next, we construct a transition \(T_i\) and store it in the experience buffer \(\mathcal{B}\). Based on the train step $ts$, we train the model whenever a certain amount of new data is added to \(\mathcal{B}\). Specifically, we employ a Temporal Difference (TD) learning strategy to train our double DQN network. First, we use the main network to compute the Q-values of the actions and select the action \(a_{j+1}^*\) with the highest Q-value. Then, the target network evaluates the chosen action's Q-value in the next state \(s_{j+1}\). This Q-value, multiplied by the discount factor \(\gamma\) and added to the current reward \(r_j\), forms the target Q-value: $q_{\text{target}} = r_j + \gamma \cdot \max_{a'} Q_{\text{target}}(s_{j+1}, a')$. The Huber Loss is used to compute the TD loss, which is formulated as:
\begin{equation}
    \mathcal{L}_{td} = 
    \begin{cases} 
    \frac{1}{2} (q - q_{\text{target}})^2, & \text{if } |q - q_{\text{target}}| < \delta \\
    \delta \cdot (|q - q_{\text{target}}| - \frac{1}{2}\delta), & \text{if } |q - q_{\text{target}}| > \delta
    \end{cases}
\end{equation}
where $\delta$ is the scale parameter, \(q\) represents the Q-value of the main network, and \(q_{\text{target}}\) denotes the Q-value of the target network. Incorporating the contrastive loss \(\mathcal{L}_c\), the overall learning objective is formulated as follows:
\begin{equation}
    \mathcal{L} = \mathcal{L}_{td} + \lambda \cdot \mathcal{L}_c
\end{equation}
where \(\lambda\) is the hyper-parameter that adjusts the weighting of the different loss functions.

\noindent \textbf{Complexity Analysis.} Our framework is designed to improve overall system efficiency by using a lightweight \textit{Decision Maker} to reduce the frequency of \textit{Predictor} invocations. Since the \textit{Decision Maker} introduces additional overhead to achieve this improvement, its efficiency must significantly surpass that of the \textit{Predictor}. Therefore, to verify this, we compare the computational complexity and number of parameters of the proposed reinforcement learning module with two ER-TTE methods \textbf{SSML}~\cite{SSML} and \textbf{MetaER-TTE}~\cite{MetaER-TTE} and a TTE methods \textbf{ConST}~\cite{ConST}. Specifically, we model the problem input as follows: the travel route comprises \( n \) road segments, the batch size is $b$, and the historical time slots are defined as $T$. We assume that the dimensions of all hidden representations are \( d \). 

Both of the two ER-TTE methods are based on \textbf{ConST}, utilizing the same embedding technique and a 3D graph attention mechanism. The embedding layer has \( O(Tnd) \) parameters, and the 3DGAT contains three dense layers, with a total parameter count of \( O(d^2) \). Following this, \textbf{ConST} employs a convolutional layer with a window size of \( w \), having \( O(wd^2) \) parameters. The complexity of computing attention scores is \( O(bn^2d) \), and the combined complexity of computing attention outputs is \( O(bn^2d + bn^2) \). Considering the complexity for each convolutional layer calculation over the entire window, it is \( O((2w+1) \cdot d \cdot d) \), and taking into account batch size and sequence length, the overall complexity is \( O(bnwd^2) \). Overall, the computational complexity of \textbf{ConST} is $O(bnwd^2 + bn^2d + bn^2)$.

\begin{table}[!t]
\centering
\caption{\revise{Complexity Analysis of the Main Methods}}
\vspace{-0.1in}
\label{tab:comp}
\scalebox{0.82}{\begin{tabular}{c|c|c}
\hline
 & \textbf{\#Params} & \textbf{Complexity} \\
\hline
\textbf{ConST} & $O(wd^2)$ & $O(bnwd^2 + bn^2d + bn^2)$ \\
\hline
\textbf{SSML} & $O(d^2)$ & $O(bn^3d + bnd^2 + bn^2d)$ \\
\hline
\textbf{MetaER-TTE} & $O((K + w)d^2)$ & $O(bnwd^2 + bn^2d + bd^2 + bKd)$ \\
\hline
\textbf{\revise{Decison Maker}} & $O(d^2)$ & $O(bn^2d + bnd^2)$ \\
\hline
\end{tabular}}
\vspace{-0.14in}
\end{table}

For \textbf{SSML}, it employs a model-based meta-learning approach. In the representation learning phase, the 3DGAT is computed for each sample in the query set and the validation set, with a complexity of \( O(n^2d) \) and a parameter count of \( O(d^2) \). For all samples in the batch, this results in a complexity of \( O(bn^3d) \). Subsequently, an MLP with complexity \( O(bnd^2) \) and an attention layer with complexity \( O(bn^2d) \) are applied. Both of these layers have parameter counts of \( O(d^2) \), and the overall complexity of \textbf{SSML} is \( O(bn^3d + bnd^2 + bn^2d) \).

For \textbf{MetaER-TTE}, an optimization-based meta-learning approach is utilized. First, the trajectory query vectors are computed with a complexity of \( O(bd^2) \), followed by the calculation of similarity scores with a complexity of \( O(bKd) \), where \( K \) represents the number of cluster centers. Next, a Cluster-aware Parameter Memory is employed, with a parameter size of \( O(Kd^2) \) and a computational complexity of \( O(d^2 + Kd) \). Finally, it propose a learning rate generator, which is a multi-layer MLP with a parameter count of \( O(d^2) \) and a computational complexity of \( O(bd^2) \). Combined with the base model, the total computational complexity is \( O(bnwd^2 + bn^2d + bd^2 + bKd) \).

As for \textbf{\modelName}, the complexity is primarily dominated by the Encoder layer. The Encoder has a parameter count of \( O(d^2) \), and its computational complexity includes the attention mechanism with \( O(bn^2d) \) and the fully connected layers with \( O(bnd^2) \). The subsequent modules also encompass these two components. Hence, the overall complexity is \( O(bn^2d + bnd^2) \).

\textbf{\textit{Remark:}} The complexity of the reinforcement learning module in \textbf{\modelName} is an order of magnitude lower than that of several baseline methods. Consequently, as demonstrated in our experiments, the efficiency of \textbf{\modelName} relative to the Predictor closely matches the model usage rate, with the computational cost of reinforcement learning constituting only a small fraction.

\section{Curriculum Learning}
\label{sec:5}

\begin{figure}
  \centering  
  \includegraphics[width=0.7\linewidth]{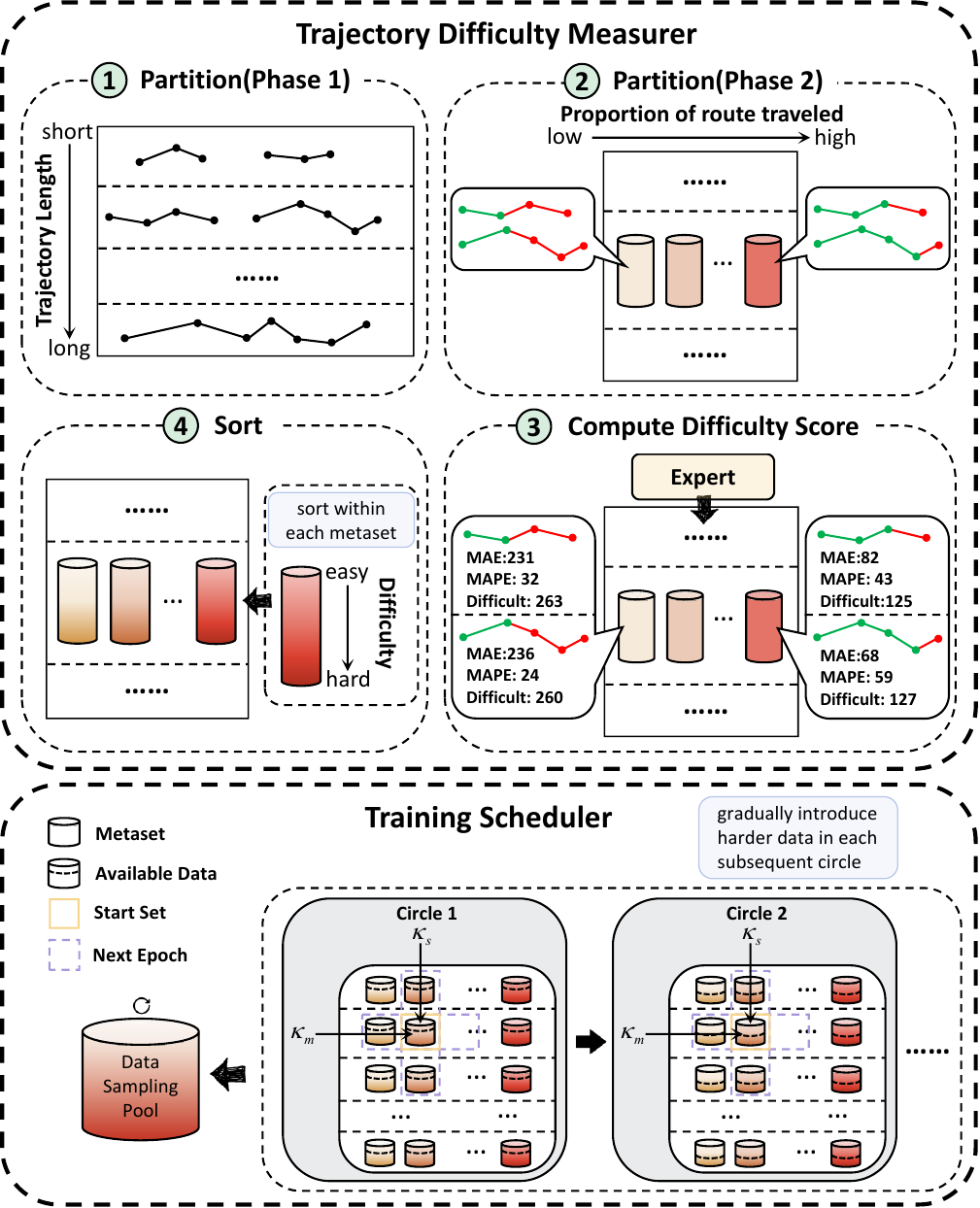}
  \caption{The Curriculum Learning Framework}
  \label{fig:cl}
\end{figure}

In our proposed end-to-end training method, each trajectory may correspond to dozens of en route requests. Each request represents a new data point due to its distinct traveled route, which leads to a significant increase in the volume of training data. Training with randomly shuffled data not only converges slowly but also tends to fall into suboptimal local minima. To mitigate this issue, we introduce curriculum learning to train the model in a meaningful sequence. Specifically, it comprises two components: the Trajectory Difficulty Measurer and the Training Scheduler. In the following sections, we will elaborate on these components in detail.

\subsection{Trajectory Difficulty Measurer} The Trajectory Difficulty Measurer evaluates the difficulty of the training samples and ranks them from easiest to hardest. The difficulty of estimating travel time can be assessed based on various factors, such as the complexity of the routes and the variability of traffic conditions. However, the model's performance on the data is of greater importance. So we use a model trained on randomly shuffled data as an expert to evaluate the difficulty of the data, employing metrics including MAE, MAPE. 

\noindent \textbf{Partition.} The direct comparison of trajectory data poses significant challenges due to the wide variation in their lengths. Firstly, trajectories of different lengths contain distinct data patterns, reflecting varying levels of prediction difficulty. For instance, shorter trajectories may contain fewer driving behaviors, providing limited useful information for the model to learn from, whereas longer trajectories encapsulate more complex and diverse driving patterns, thereby substantially increasing the difficulty of accurate prediction. Secondly, we rely on the expert to evaluate trajectory difficulty using multiple metrics, however, these metrics vary considerably for trajectories of different remaining route lengths.


Given the aforementioned reasons, we argue that prior to conducting trajectory difficulty assessment, it is essential to partition the trajectory data. Hence, as illustrated in Fig.~\ref{fig:cl}, we partition the trajectories in two phases, and separately group the data according to two factors: the route length and the proportion of the route traveled. First, we partition the dataset \(D\) based on the length of the route into \(N\) disjoint subsets, i.e., \(D = \{\tilde{D}_1, \tilde{D}_2, \ldots, \tilde{D}_N\}\), where \(\tilde{D}_i\) represents the \(i\)-th subset and \(\tilde{D}_i \cap \tilde{D}_j = \emptyset, \forall i \neq j, i, j \in [1, N]\). Each subset is then further divided into \(M\) metasets based on the proportion of the route traveled, i.e., \( D = \{ \tilde{D}_{ij} \mid i = 1, 2, \ldots, N; \; j = 1, 2, \ldots, M \} \). For clarity, in the following text, we refer to \(\tilde{D}_{i}\) as a "subset" and \(\tilde{D}_{ij}\) as a "metaset".

\noindent \textbf{Difficulty Assessment.} After partitioning the trajectory data, difficulty assessment only needs to be conducted within each individual metaset. The difficulty score is defined as follows: 
\begin{equation}
\mu_i = \text{MAE}(\mathcal{M}_e(\mathbf{x}_i), \mathbf{y}_i) + \text{MAPE}(\mathcal{M}_e(\mathbf{x}_i), \mathbf{y}_i)
\end{equation}
where \(\mathbf{x}_i\) is the input, \(\mathbf{y}_i\) is the ground truth, and \(\mathcal{M}_e\) denotes the expert. In Fig.~\ref{fig:cl}, we provide illustrative examples of the corresponding difficulty scores for trajectories within different metasets. MAE provides a direct reflection of the absolute error in expert evaluations, while MAPE captures relative error, offering a better measure of the prediction’s accuracy in relation to the true values. By combining both MAE and MAPE, we address the limitations of relying on a single metric, allowing for a more comprehensive assessment of trajectory difficulty across different magnitudes and scales. We then sort the trajectories based on their difficulty scores.

\begin{figure}[!t]
\begin{algorithm}[H]
    \small 
    \SetNlSty{textbf}{}{}
    \SetKwInOut{Input}{Input}
    \SetKwInOut{Output}{Output}
    
    \Input{training dataset $\mathbf{D}$, subset number $\mathbf{N}$, metaset number $\mathbf{M}$, the Predictor $\mathcal{M}_{p}$, start position $\kappa_s$ and $\kappa_m$, training circle $cr$, training epochs $ep$, tolerance for early stopping $\tau_{tol}$.}
    \Output{parameters $\theta_{p}$, for the Predictor $\mathcal{M}_{p}$.}
    \renewcommand{\thealgocf}{2} 
    
    \caption{Training Scheduler for \textbf{\modelName}}
    \label{alg:ts}
    initialize $\theta_{p}$ with normal distribution\;
    \For{$m \gets 1 \ldots cr$}{
        release the more challenging data within each metaset\;
        Add $\tilde{D}_{\kappa_s\kappa_m}$ to sampling pool\;
        \For{$i \gets 1 \ldots ep$}{
            $\theta_{p} \gets$ train the Predictor with the sampling pool\;
            update the Predictor with $\theta_{p}$\;
            \If{$i < \max\left( N - \kappa_s,\ \kappa_s - 1,\ M - \kappa_m,\ \kappa_m - 1 \right) + 1$}{
                include the neighboring sets of all newly added sets\;
            }
            calculate validation loss $L_{val}$\;
            \If{$|L_{val}^{i} - L_{val}^{i-1}| < \tau_{tol}$}{
                break\;
            }
        }
    }
\end{algorithm}
\end{figure}

\subsection{Training Scheduler}
Since we have already ordered the data within each metaset, an important task of the Training Scheduler is to determine how to arrange the metaset \(\tilde{D}_{ij}\) for training. A straightforward approach is to directly select the simpler data from all metasets and add it to the sampling pool. However, as we mentioned earlier, this approach overlooks the varying difficulty levels among different metasets. Therefore, we propose that the data from different metasets should be sequentially introduced into the sampling pool throughout the curriculum. Intuitively, we believe that training should begin with datasets of moderate length and a balanced proportion of the traveled route, as their driving behavior patterns are easier to learn. Additionally, starting with medium-length trajectories facilitates generalization to both shorter and longer routes.

As presented in Algorithm~\ref{alg:ts}, we design a dual-loop training method. This approach initiates with the training circle, an outer loop that iterates through all metasets, starting by training on easier data drawn from these sets. As each subsequent cycle progresses, more challenging data is gradually introduced. Within each major cycle, we define two hyperparameters, $\kappa_s$ and $\kappa_m$, to determine the starting position of the metaset for training, denoted as \(\tilde{D}_{\kappa_s\kappa_m}\). As illustrated in Figure~\ref{fig:cl}, $\kappa_s$ is used to select a subset, while $\kappa_m$ specifies the position of this metaset within this subset. Due to the unsuitability of edge positions in the metaset matrix for initiating training, the values of these two parameters are chosen from a moderate range of positions. 

The inner loop is defined by epochs. At the start of each epoch, data is randomly sampled from the Data Sampling Pool for training. After each training session, we expand the Sampling Pool by adding any neighboring metasets of the newly introduced metaset that have not yet been included. For example, if (\(\tilde{D}_{ij}\)) is added, we also include its neighbors (\(\tilde{D}_{i \pm 1, j \pm 1}\)). This process continues until all metasets are incorporated into the Sampling Pool. Finally, we apply an early stopping strategy—once the model has been trained sufficiently on the current data, we conclude the major cycle and move on to the next one.

In the specific scenario of ER-TTE, we first partition the dataset into multiple metasets to effectively manage the training data. Then, we evaluate the difficulty of the data from two perspectives: the length of the trajectories and the starting locations of requests along the routes, as well as the complexity of trajectory data for prediction using deep learning models. In the subsequent experimental section, we will thoroughly investigate the impact of curriculum learning parameters on the results.

\section{EXPERIMENTS}
\label{sec:6}

In this section, we evaluate the effectiveness, efficiency, and scalability of our method \textbf{\modelName} on three real-world datasets.

\subsection{Experimental Setup}

\begin{table}[h]
\centering
\caption{\revise{Statistics of Datasets}}
\label{tab:dataset}
\scalebox{0.75}{\begin{tabular}{c|ccc}
\hline
 & \textbf{Chengdu} & \textbf{Xi'an} & \textbf{\revise{Porto}} \\
\hline
Travel routes & 5.8M & 3.4M & \revise{0.5M}\\
\hline
Avg travel time (s) & 500.65 & 757.07 & \revise{479.12}\\
\hline
Time interval & 10/1--11/30, 2016 & 10/1--11/30, 2016 & \revise{7/1, 2013--6/30, 2014} \\
\hline
Avg \# of points & 180 & 205 & \revise{36}\\
\hline
Avg \# of road segments & 17 & 25 & \revise{19}\\
\hline
Avg length (meter) & 3,477.85 & 4,143.17 & \revise{2471.38}\\
\hline
\end{tabular}}
\end{table}

\noindent \textbf{Datasets.} (1) \textbf{Road Networks.} We utilize three road networks in our experiments: \textit{Chengdu Road Network (CRN)}, \textit{Xi'an Road Network (XRN)}, and \textit{Porto Road Network (PRN)}, which are all extracted from OpenStreetMap~\cite{OpenStreetMap}. \textit{CRN} includes 3191 vertices and 9468 edges; \textit{XRN} contains 4576 vertices and 12668 edges; \textit{PRN} contains 2755 vertices and 5315 edges.

\noindent (2) \textbf{Travel Route.} The trajectory data for \textit{Chengdu} and \textit{Xi'an} were collected from 10/01/2016 to 11/30/2016, while the trajectory data for \textit{Porto} were collected from 07/01/2013 to 06/30/2014. The \textit{Chengdu} dataset contains 5.8 million trajectories, the \textit{Xi'an} dataset comprises 3.4 million trajectories, and the \textit{Porto} dataset includes 0.5 million trajectories. We apply a map matching algorithm to align the GPS trajectories with the corresponding road networks, thereby deriving the travel routes for the three datasets. Table~\ref{tab:dataset} presents the statistics of these datasets, where \textit{Avg \# of points} indicates the average number of GPS points in each travel route, and \textit{Avg \# of road segments} represents the average number of road segments traversed by each travel route.

\noindent (3) \textbf{Traffic Condition.} To calculate the historical traffic conditions, we partition the trajectory data into time slots \(\Delta T\), where \(\Delta T\) is set to 5 minutes in our work. We compute various traffic speeds for each road segment under different time slots, such as the mean and median values. For each travel route, we consider \(p\) past time slots as the traffic conditions for that trajectory.

\noindent (4) \textbf{Background Information.} We collect weather conditions and compile data on holidays, weekends, and rush-hour as background information. Weather information is gathered from the website~\cite{weather}, and the weather conditions are categorized into 16 types.

\noindent (5) \textbf{Training, Validation and Test.} We partition the datasets based on the day of data collection. Specifically, the datasets are divided into training, validation, and test sets according to the ratios of 70\%:10\%:20\%, respectively.

\noindent \textbf{Baseline methods.}
We compare our models with six baseline methods:

\vspace{-0.06in}

\begin{itemize}[leftmargin=10pt]
    \item \textbf{Avg}: We accumulate historical trajectory data for each time slot on every link, and compute the average speed within that time slot. The computed averages across all time slots for each link are then used to predict the travel time.
    \item \textbf{DeepTTE}~\cite{DeepTTE}: This is a spatio-temporal neural network. It first encodes the spatial information through Geo-Conv, then encodes the temporal information using LSTM, and employs multi-task learning to predict the travel time for the entire route.
    \item \textbf{STANN}~\cite{STANN}: This is a spatio-temporal attentive neural network. It proposes an LSTM-based encoder–decoder architecture to explore spatio-temporal relationships and introduces an attention mechanism in the decoder.
    \item \textbf{ConST}~\cite{ConST}: This is a spatio-temporal graph neural network. It proposes a graph attention mechanism to capture the spatial and temporal dependencies to estimate the travel time.
    \item \textbf{SSML}~\cite{SSML}: This is a meta-learning model for ER-TTE. It constructs ER-TTE as a few-shot learning problem and proposes a model-based self-supervised meta-learning method to learn meta-knowledge for travel time estimation.
    \item \textbf{MetaER-TTE}~\cite{MetaER-TTE}: This is a state-of-the-art meta-learning model for ER-TTE. It proposes an adaptive meta-learning model based on MAML~\cite{MAML}, providing different initialization parameters and learning rates for trajectories with various contextual information.
\end{itemize}

\noindent \textbf{End-to-end Evaluation.} Unlike previous approaches~\cite{SSML, MetaER-TTE} that use a fixed proportion of traveled distance for offline model training and evaluation, we propose an end-to-end evaluation method. In this method, mirroring real-world online scenarios, prediction requests are initiated at regular intervals along each trajectory. For each travel route, the metrics corresponding to all requests are averaged to provide an overall evaluation of the model’s performance throughout the entire route.

\noindent \textbf{Evaluation Metrics.} We evaluate our proposed methods and baseline methods based on three metrics: RMSE (Root Mean Square Error), MAE (Mean Absolute Error) and MAPE (Mean Absolute Percent Error), which are widely used by the baselines we compare with. Specifically, suppose that the ground truth of the remaining route is represented as $\mathbf{y} = \{y_{r}^i\}$ and the predicted travel time is denoted as $\mathbf{\hat{y}} = \{\hat{y}_{r}^i\}$, these metrics are defined as: $RMSE(\mathbf{y}, \mathbf{\hat{y}}) = \sqrt{\frac{1}{N} \sum_{i=1}^N (y_{r}^i - \hat{y}_{r}^i)^2}$; $MAE(\mathbf{y}, \mathbf{\hat{y}}) = \frac{1}{N} \sum_{i=1}^N \left| y_{r}^i - \hat{y}_{r}^i \right|$; $MAPE(\mathbf{y}, \mathbf{\hat{y}}) = \frac{1}{N} \sum_{i=1}^N \left| \frac{y_{r}^i - \hat{y}_{r}^i}{y_{r}^i} \right|$,
where $N$ denotes the number of the requests included in a travel route. We calculate the average of these metrics across all travel routes to evaluate the performance of all methods. In addition to the three base performance metrics, we also compared an efficiency metric: MUR (Model Utilization Rate), defined as the proportion of prediction requests that required recalculation by the Predictor.

\noindent \textbf{Experimental Settings.} All deep learning methods were implemented with PyTorch 2.0.1 and Python 3.11.5, and trained with a Tesla V100 GPU. The platform ran on Ubuntu 18.04 OS. In addition, we used Adam~\cite{adam} as the optimization method with the mini-batch size of 512. The learning rate is set as 0.0001, and the training epoch is set as 100, and an early stopping mechanism is adopted. For the Double DQN, the experience replay buffer size is set to 500,000, the target network update step to 2,000, and the discount factor to 0.9.

\begin{figure}
  \centering  
  \includegraphics[width=0.8\linewidth]{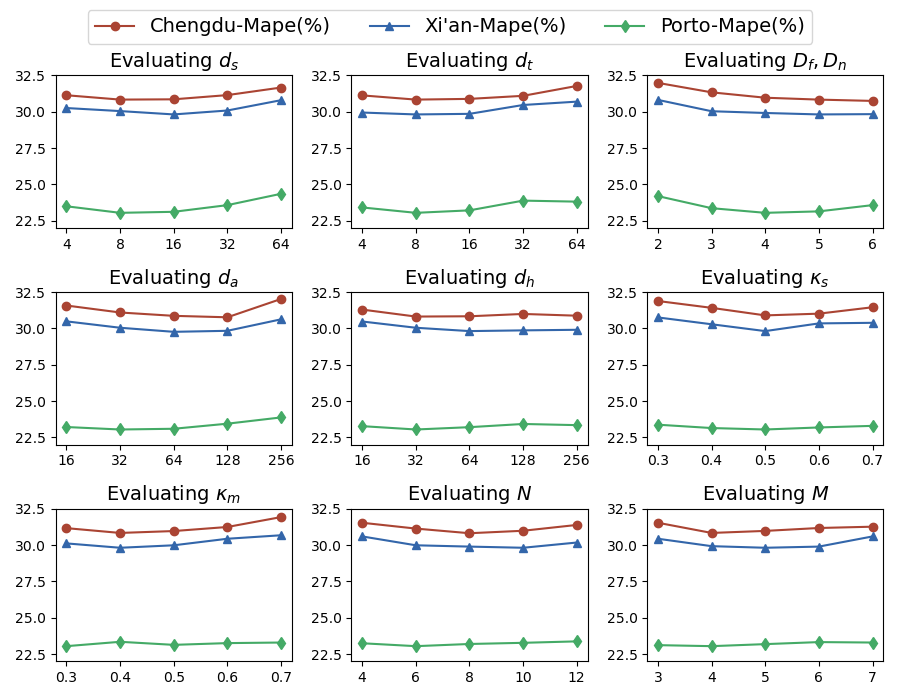}
  \vspace{-0.15in}
  \caption{MAPE(\%) vs. Hyper-parameters}
  \vspace{-0.15in}
  \label{fig:hype}
\end{figure}

\subsection{Setting of Model's Hyper-parameters}
we consider the following hyper-parameters: (1) the embedding size ($d_s$, $d_t$) of road segments and time slots; (2) the depth ($D_f$, $D_n$) of the Offline Feature Encoder and Online Feature Encoder; (3) the settable-dimension size ($d_a$) of attention mechanism; (4) the settable-dimension size ($d_h$) of certain hidden representations; (5) the starting positions ($\kappa_s$, $\kappa_m$) for curriculum learning; (6) the partition sizes of the subset and metaset ($N$, $M$). In particular, given a hyper-parameter, we first select its value range according to the experience under some constraints (e.g., the limitation of GPU memory). Then, we conduct experiments on the validation \textit{Chengdu}, \textit{Xi'an} and \textit{Porto} to determine its optimal value. As shown in Fig.~\ref{fig:hype}, we plot the MAPE for different hyper-parameters. In summary, we set each hyper-parameter with the value corresponding to the optimal performance as follows: (1) For \textit{Chengdu}, we have $d_s = 8$, $d_t = 8$, $D_f = 6$, $D_n = 6$, $d_a = 128$, $d_h = 32$, $\kappa_s = 0.5$, $\kappa_m = 0.4$, $N = 8$, $M = 4$. (2) For \textit{Xi'an}, we have $d_s = 16$, $d_t = 8$, $D_f = 5$, $D_n = 5$, $d_a = 64$, $d_h = 64$, $\kappa_s = 0.5$, $\kappa_m = 0.4$, $N = 10$, $M = 5$. (3) For \textit{Porto}, we have $d_s = 8$, $d_t = 8$, $D_f = 4$, $D_n = 4$, $d_a = 32$, $d_h = 32$, $\kappa_s = 0.5$, $\kappa_m = 0.3$, $N = 6$, $M = 4$.

\begin{figure}
  \centering  
  \includegraphics[width=0.85\linewidth]{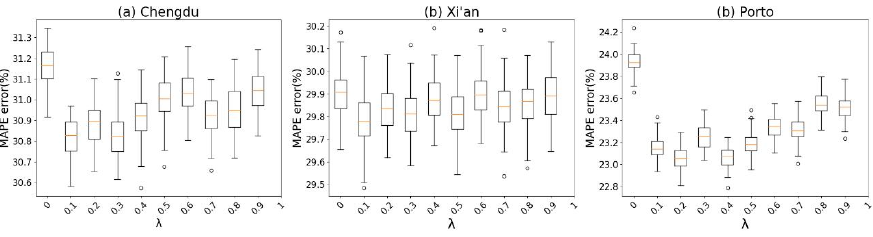}
  \vspace{-0.2in}
  \caption{MAPE(\%) vs. Loss Weight \( \lambda \)}
  \vspace{-0.1in}
  \label{fig:loss}
\end{figure}

\subsection{Effectiveness of Loss Weight}
To fine-tune the loss weight \( \lambda \), we vary it from 0 to 1 in increments of 0.1 during the training of \textbf{\modelName}. We compute the MAPE metric on the validation data and plot the corresponding box plots for the results of the three datasets in Figure~\ref{fig:loss}. Our analysis indicates that performance significantly improves as \(\lambda\) varies from 0 to 0.1, demonstrating the effectiveness of the contrastive learning module. However, as \(\lambda\) increases further, the model's performance generally exhibits a downward trend, indicating that the reinforcement learning loss becomes more critical and dominant. According to the majority voting rule, the optimal values of \(\lambda\) for \textit{Chengdu}, \textit{Xi'an}, and \textit{Porto} are 0.1, 0.1, and 0.2, respectively, which are set as the default values in subsequent experiments.

\begin{table*}[ht]
\centering
\caption{\revise{Effectiveness Results on Test Data.}}
\vspace{-0.1in}
\label{tab:performance}
\scalebox{0.59}{
\begin{tabular}{c|cccc|cccc|cccc}
  \hline
  {Dataset} & \multicolumn{4}{c|}{Chengdu} & \multicolumn{4}{c|}{Xi'an} & \multicolumn{4}{c}{\revise{Porto}} \\
  \hline
  {Method} & \textbf{MAE (s)} & \textbf{RMSE (s)} & \textbf{MAPE (\%)} & \textbf{MUR (\%)} & \textbf{MAE (s)} & \textbf{RMSE (s)} & \textbf{MAPE (\%)} & \textbf{MUR (\%)} & \textbf{\revise{MAE (s)}} & \textbf{\revise{RMSE (s)}} & \textbf{\revise{MAPE (\%)}} & \textbf{\revise{MUR (\%)}}\\
  \hline
  \textbf{Avg} & 131.93 & 190.14 & 47.35 & -- & 139.28 & 204.58 & 44.12 & -- & \revise{91.32} & \revise{129.96} & \revise{39.64} & -- \\
  \hline
  \textbf{DeepTTE} & 120.78 & 167.85 & 43.15 & 100\% & 128.51 & 176.11 & 39.47 & 100\% & \revise{69.92} & \revise{94.09} & \revise{29.86} & \revise{100\%} \\
  \hline
  \textbf{STANN} & 118.35 & 161.74 & 42.92 & 100\% & 129.67 & 174.63 & 39.86 & 100\% & \revise{71.34} & \revise{100.94} & \revise{30.71} & \revise{100\%} \\
  \hline
  \textbf{ConST} & 112.36 & 143.18 & 41.26 & 100\% & 122.79 & 156.31 & 38.79 & 100\% & \revise{68.16} & \revise{92.07} & \revise{28.83} & \revise{100\%} \\
  \hline
  \textbf{SSML} & 94.81 & 126.15 & 34.46 & 100\% & 106.43 & 140.89 & 33.02 & 100\% & \revise{60.27} & \revise{79.15} & \revise{25.74} & \revise{100\%} \\
  \hline
  \textbf{MetaER-TTE} & 90.74 & 114.69 & 33.87 & 100\% & 103.81 & 134.09 & 31.49 & 100\% & \revise{58.30} & \revise{73.23} & \revise{24.91} & \revise{100\%} \\
  \hline
  \hline
  \textbf{ND}& 89.39 & 112.51 & 33.09 & 34.58\% & 101.38 & 129.97 & 30.94 & 35.44\% & \revise{59.42} & \revise{80.35} & \revise{25.80} & \revise{35.92\%} \\
  \hline
  \textbf{NU} & 86.17 & 108.46 & 31.61 & 24.33\% & 98.23 & 125.61 & 30.42 & 25.67\% & \revise{54.77} & \revise{70.84} & \revise{23.27} & \revise{23.55\%} \\
  \hline
  \textbf{NA} & 85.32 & 105.93 & 31.17 & 26.05\% & 96.97 & 122.49 & 29.87 & 26.71\% & \revise{55.26} & \revise{71.39} & \revise{23.94} & \revise{20.39\%} \\
  \hline
  \hline
  \revise{\textbf{\modelName-s}} & 90.83 & 115.92 & 33.79 & 29.64\% & 101.94 & 131.29 & 31.27 & 32.18\% & \revise{56.93} & \revise{75.87} & \revise{24.16} & \revise{27.86\%} \\
  \hline
  \revise{\textbf{\modelName-p}} & \revise{87.58} & \revise{111.07} & \revise{32.45} & \revise{36.01\%} & \revise{100.72} & \revise{128.26} & \revise{31.05} & \revise{31.59\%} & \revise{57.61} & \revise{77.21} & \revise{25.06} & \revise{28.19\%} \\
  \hline
  \revise{\textbf{\modelName-a}} & \revise{91.81} & \revise{121.73} & \revise{34.38} & \revise{42.76\%} & \revise{101.47} & \revise{136.52} & \revise{31.44} & \revise{39.62\%} & \revise{56.74} & \revise{74.65} & \revise{24.53} & \revise{34.02\%} \\
  \hline
  \textbf{\modelName} & \textbf{84.76} & \textbf{103.75} & \textbf{30.83} & \textbf{25.58\%} & \textbf{96.59} & \textbf{121.68} & \textbf{29.81} & \textbf{24.36\%} & \revise{\textbf{54.13}} & \revise{\textbf{68.78}} & \revise{\textbf{23.04}} & \revise{\textbf{22.81\%}} \\
  \hline
  
\end{tabular}}

\end{table*}

\subsection{Effectiveness Comparison}
Apart from comparing \textbf{\modelName} with baseline methods, we also propose three ablation study variants. In \textbf{ND}, we replace our proposed Double DQN network with a Transformer~\cite{transformer}. All features are concatenated and then fed directly into the Transformer for decision-making. In \textbf{NU}, we remove the curriculum learning strategy and train the \textit{Predictor} using all end-to-end data in a randomly shuffled order. In \textbf{NA}, we remove the contrastive loss and directly merge the online and offline features. To thoroughly investigate the role of the reinforcement learning agent in the framework, we further introduce agents based on the PPO~\cite{PPO} method and the A3C~\cite{A3C} method, integrated with the model architecture proposed in this paper, and denote them as \textbf{\modelName-p} and \textbf{\modelName-a}, respectively. All the aforementioned experiments use \textbf{MetaER-TTE} as the \textit{Predictor}. Additionally, in \textbf{\modelName-s}, we use \textbf{SSML} as the \textit{Predictor} to demonstrate the robustness of our framework. 

Table~\ref{tab:performance} reports the performance of all methods using different metrics on the three datasets, and we have the following observations:

\noindent (1) In end-to-end evaluation, the ER-TTE method demonstrates a significant performance improvement compared to the traditional TTE method, as its performance gradually enhances with the increase in the traveled distance. However, all methods exhibit relatively high MAPE values. This is attributed to the shortening of the remaining distance, which reduces the length of the trajectory to be predicted and consequently increases the relative percentage error.

\noindent (2) \textbf{Avg} performs worse compared to deep learning methods, because the former can approximately fit any function. Moreover, when historical data are sparse, the \textbf{Avg} method is insufficient for providing accurate predictions.

\noindent (3) In analyzing the results of the ablation experiments on \textbf{ND}, \textbf{NU}, and \textbf{NA}, we observed that the removal of the Double DQN network had the most significant impact, with a notable increase in both effective metrics and the MUR metric. This indicates that only using a single Transformer for decision-making is insufficient to capture the complex spatio-temporal correlation in online scenarios. The experiments on \textbf{NU} and \textbf{NA} demonstrate the effectiveness of our curriculum learning and contrastive learning components. While their impact on MUR was smaller, they still contributed to performance improvements.

\noindent (4) Comparing the experimental results of \textbf{\modelName-s} and \textbf{SSML}, we find that our proposed framework is effective across multiple ER-TTE predictors, thereby demonstrating the robustness of our approach.

\noindent (5) The experimental results of \textbf{\modelName-p} and \textbf{\modelName-a} demonstrate that the Double DQN method exhibits superior performance in this scenario. This is primarily because Double DQN is better suited for reinforcement learning tasks that involve controlling discrete action spaces, especially when the number of actions is limited. In contrast, methods such as PPO and A3C are more appropriate for continuous action spaces. Given the relatively small action space in this task, the value-based Double DQN method can quickly learn the optimal Q-value for each action. While policy gradient-based methods, such as PPO and A3C, often result in overly complex policy optimization steps, which are more prone to instability and overfitting.

\noindent (6) \textbf{\modelName} outperforms all other models across all metrics. This is attributed to our method's analysis of driving behavior patterns, which selectively inputs beneficial real-time information into the \textit{Predictor}, rather than feeding all newly observed data into the \textit{Predictor} indiscriminately.

\noindent (7) The results of the MUR metric indicate that in real-world scenarios, the majority of real-time prediction requests do not necessitate re-computation by the estimated model. Compared to existing methods, our approach reduces the computational load by approximately fourfold, significantly enhancing the efficiency.

\noindent (8) When comparing the errors on the \textit{Chengdu} and \textit{Xi'an} datasets, all methods exhibit lower MAE and RMSE metrics on the \textit{Chengdu} dataset, while the MAPE metric is relatively higher. This is primarily due to the shorter trajectories in the \textit{Chengdu} dataset compared to those in \textit{Xi'an}, resulting in smaller absolute errors but larger percentage errors when normalized by the overall trajectory length.

\noindent (9) The metrics on the \textit{Porto} dataset are lower than \textit{Chengdu} and \textit{Xi'an} datasets. This is because the trajectories in the \textit{Porto} dataset are shorter, and the GPS sampling frequency is lower, resulting in coarser granularity. These characteristics reduce the prediction difficulty for this dataset.

\begin{figure}
  \centering  
  \includegraphics[width=0.85\linewidth]{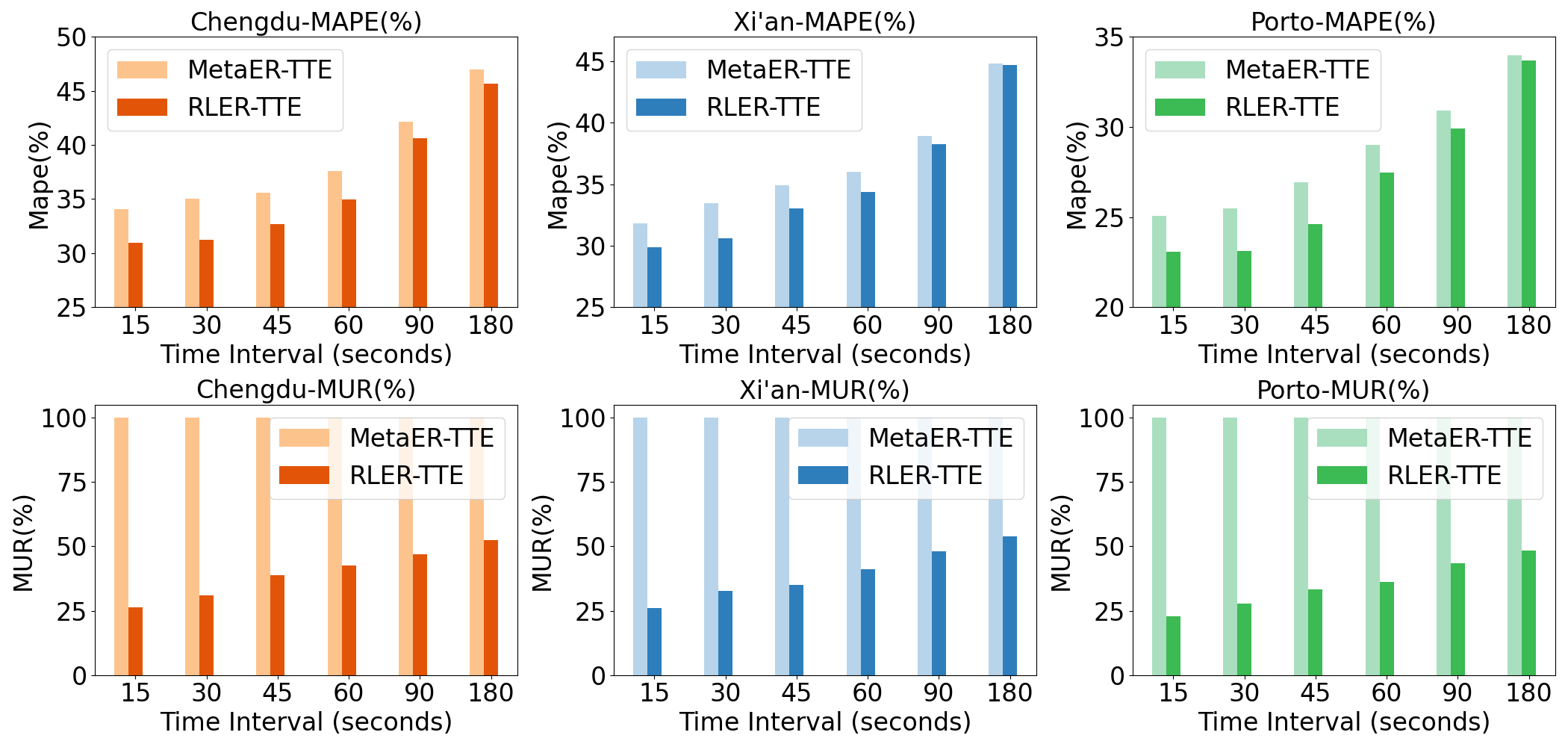}
  \caption{MAPE(\%), MUR(\%) vs. Time Interval}
  \label{fig:intv}
\end{figure}

\subsection{Effectiveness of Request Interval}
The interval $\Delta t$ between real-time requests is a critical parameter for the online ER-TTE task, which has not been explored in previous work~\cite{SSML, MetaER-TTE}. To investigate the performance of our method under different values of $\Delta t$, we compare the MAPE and MUR metrics of \textbf{\modelName} with the state-of-the-art method \textbf{MetaER-TTE} across various values of $\Delta t$, as shown in Figure~\ref{fig:intv}. We summarize our findings as follows:

\noindent (1) As the time interval $\Delta t$ increases, the MAPE values of both methods rise accordingly. This indicates that more frequent en-route predictions lead to higher prediction accuracy.

\noindent (2) Our method demonstrates a consistent advantage over the baseline in terms of the MAPE metric across all intervals. However, as the prediction interval increases, the performance gains gradually diminish. This is due to the interval becoming excessively long relative to the trajectory. For instance, the average travel times in the \textit{Chengdu} and \textit{Porto} datasets are 500.65s and 479.12s, respectively. As the prediction interval increases, the number of en-route predictions decreases significantly, which is inconsistent with the application scenario of ER-TTE techniques and does not meet the requirements of real-world deployment systems.

\noindent (3) The MUR metric of \textbf{\modelName} increases as $\Delta t$ grows, since a longer time interval implies that the newly collected real-time information at each prediction step is more valuable. Additionally, our frequency reward encourages the system to perform re-prediction when the duration of the previous prediction interval becomes longer. However, compared to the baseline method, our approach consistently maintains a significant improvement in the MUR metric, which greatly enhances efficiency.

\begin{table}
\centering
\caption{\revise{Efficiency of Different Methods}}
\vspace{-0.15in}
\label{tab:efficiency}
\scalebox{0.8}{%
\begin{tabular}{c|ccc|ccc|ccc}
  \hline
  \multirow{3}{*}{Method} & \multicolumn{3}{c|}{memory size} & \multicolumn{3}{c|}{training time} & \multicolumn{3}{c}{estimation latency}\\
  {} & \multicolumn{3}{c|}{(MByte)} & \multicolumn{3}{c|}{(minutes/ep)} & \multicolumn{3}{c}{(seconds/K)}\\
  \cline{2-10}
  {} & Chengdu & Xi'an & \revise{Porto} & Chengdu & Xi'an & \revise{Porto} & Chengdu & Xi'an & \revise{Porto} \\
  \hline
  \textbf{Avg} & 183MB & 126MB & \revise{21MB} & - & - & - & - & - & - \\
  \textbf{DeepTTE} & 3.8MB & 3.9MB & \revise{3.6MB} & 14.47 & 9.72 & \revise{1.88} & 1.27s & 1.53s & \revise{1.31s} \\
  \textbf{STANN} & 6.4MB & 6.5MB & \revise{6.1MB} & 5.59 & 3.41 & \revise{0.73} & 0.61s & 0.88s & \revise{0.65s} \\
  \textbf{ConST} & 5.9MB & 6.9MB & \revise{5.8MB} & 10.46 & 6.74 & \revise{1.15} & 0.69s & 0.93s & \revise{0.71s} \\
  \textbf{SSML} & 6.2MB & 7.4MB & \revise{5.9MB} & 16.22 & 12.68 & \revise{2.07} & 1.19s & 1.48s & \revise{1.26s} \\
  \textbf{MetaER-TTE} & 7.1MB & 8.0MB & \revise{6.4MB} & 13.34 & 8.41 & \revise{1.35} & 0.78s & 1.04s & \revise{0.86s} \\
  \revise{\textbf{ND}} & \revise{5.7MB} & \revise{6.8MB} & \revise{5.1MB} & \revise{7.91} & \revise{4.83} & \revise{0.82} & \revise{0.35s} & \revise{0.44s} & \revise{0.38s} \\
  \revise{\textbf{NU}} & \revise{6.8MB} & \revise{8.9MB} & \revise{6.4MB} & \revise{11.35} & \revise{7.29} & \revise{1.16} & \revise{0.29s} & \revise{0.38s} & \revise{0.28s} \\
  \revise{\textbf{NA}} & \revise{6.8MB} & \revise{8.8MB} & \revise{6.3MB} & \revise{11.24} & \revise{7.08} & \revise{1.13} & \revise{0.30s} & \revise{0.39s} & \revise{0.27s} \\
  \textbf{\modelName-s} & 6.9MB & 9.1MB & \revise{6.4MB} & 12.01 & 7.94 & \revise{1.28} & 0.45s & 0.59s & \revise{0.46s} \\
  \revise{\textbf{\modelName-p}} & \revise{9.1MB} & \revise{13.6MB} & \revise{6.8MB} & \revise{10.39} & \revise{6.18} & \revise{1.07} & \revise{0.39s} & \revise{0.45s} & \revise{0.40s} \\
  \revise{\textbf{\modelName-a}} & \revise{9.1MB} & \revise{11.3MB} & \revise{6.8MB} & \revise{9.36} & \revise{5.94} & \revise{1.02} & \revise{0.44s} & \revise{0.56s} & \revise{0.42s} \\
  \textbf{\modelName} & 6.8MB & 8.9MB & \revise{6.4MB} & 11.47 & 7.22 & \revise{1.17} & \textbf{0.29s} & \textbf{0.37s} & \revise{\textbf{0.27s}} \\
  \hline
\end{tabular}}
\vspace{-0.1in}
\end{table}

\subsection{Efficiency Comparison}
We compare efficiency in terms of memory usage, training time, and estimation latency. Memory usage measures the amount of memory consumed by each method to evaluate memory efficiency. Training time is used to compare the offline learning efficiency of deep learning-based methods, calculated as the average time per epoch for each method. Estimation latency assesses the efficiency of applying these methods in online scenarios. We selected 1,000 routes and evaluated the time required to perform complete real-time predictions with multiple real-time requests on these trajectories. The results are reported in Table~\ref{tab:efficiency}. Our observations are as follows:

\noindent (1) The \textbf{Avg} method requires more memory compared to other approaches, as it needs to load data proportional to the size of the trajectory history, whereas other methods only need to load the model parameters.

\noindent (2) Due to the advantages of the attention mechanism, the training and estimation efficiency of \textbf{STANN} and \textbf{ConST} is higher. Meanwhile, the meta-learning components of \textbf{SSML} and \textbf{MetaER-TTE} contribute to an increased computational cost for these methods.

\noindent (3) Comparing the datasets of \textit{Xi'an}, \textit{Chengdu} and \textit{Porto}, it is observed that the model parameters of the \textit{Xi'an} dataset are more numerous. This is attributed to the larger road network in \textit{Xi'an}. In contrast, the training time for the \textit{Chengdu} dataset is longer, while that of \textit{Porto} is much shorter due to differences in dataset sizes. Regarding estimation latency, the dataset from \textit{Xi'an} requires a longer inference time compared to both \textit{Chengdu} and \textit{Porto}. This is because the trajectory lengths in \textit{Xi'an} are generally longer.

\noindent (4) Our approach, \textbf{\modelName}, achieves significantly better efficiency compared to prevailing state-of-the-art methods. This enhancement is primarily attributed to the efficiency of the lightweight reinforcement learning model, with the improvement ratio closely aligning with the model utilization rate of the \textit{Predictor} within the framework.

\noindent (5) Comparing the results of \textbf{ND}, \textbf{NU}, and \textbf{NA}, we observe that the \textbf{ND} model requires fewer parameters and has a shorter training time due to its simpler architecture. On the other hand, \textbf{NU} and \textbf{NA} have a minimal impact on efficiency, as the system’s efficiency is primarily depends on the MUR metric. The primary purpose of these two components is to enhance the system’s performance, without compromising the framework’s inherent efficiency. The experimental results of \textbf{\modelName-p} and \textbf{\modelName-a} show that they consume more memory but deliver relatively inferior efficiency, further demonstrating that the double DQN method is better suited for the application scenario of our framework.

\noindent (6) The experimental results on \textbf{\modelName-s} demonstrate that our framework possesses plug-and-play characteristics, allowing it to adapt to various estimated models. This robust generalization capability motivates us to explore more complex ER-TTE estimation methods. Moreover, more accurate models imply lower MUR, and with the aid of our framework, the efficiency cost can be significantly reduced.


\begin{table}[h]
\centering
\vspace{-0.1in}
\caption{\revise{Scalability of methods, measured by MAPE(\%)}}
\vspace{-0.1in}
\label{tab:sca}
\scalebox{0.8}{\begin{tabular}{c|ccccc}
\hline
\revise{Dataset} & \multicolumn{5}{c}{\textbf{\revise{Chengdu/Xi'an}}} \\
\hline
\revise{Scale} & \revise{20\%} & \revise{40\%} & \revise{60\%} & \revise{80\%} & \revise{100\%} \\
\hline
\textbf{\revise{Avg}} & \revise{58.78/55.34} & \revise{52.43/49.15} & \revise{49.12/46.03} & \revise{48.12/44.89} & \revise{47.35/44.12} \\
\hline
\textbf{\revise{DeepTTE}} & \revise{44.10/41.73} & \revise{44.38/40.98} & \revise{43.87/40.38} & \revise{43.59/39.91} & \revise{43.15/39.37} \\
\hline
\textbf{\revise{STANN}} & \revise{44.82/41.76} & \revise{44.12/41.35} & \revise{43.78/40.61} & \revise{43.29/40.23} & \revise{42.92/39.86} \\
\hline
\textbf{\revise{ConST}} & \revise{43.49/41.73} & \revise{43.14/41.01} & \revise{42.39/40.23} & \revise{41.59/39.18} & \revise{41.26/38.79} \\
\hline
\textbf{\revise{SSML}} & \revise{36.98/35.99} & \revise{35.79/34.83} & \revise{35.21/34.12} & \revise{34.76/33.48} & \revise{34.46/33.02} \\
\hline
\textbf{\revise{MetaER-TTE}} & \revise{\underline{37.23}/\underline{35.61}} & \revise{\underline{35.41}/\underline{33.76}} & \revise{\underline{34.52}/\underline{32.72}} & \revise{\underline{34.01}/\underline{31.83}} & \revise{\underline{33.47}/\underline{31.49}} \\
\hline
\textbf{\revise{RLER-TTE}} & \revise{\textbf{33.97}/\textbf{33.41}} & \revise{\textbf{32.64}/\textbf{31.85}} & \revise{\textbf{31.98}/\textbf{30.97}} & \revise{\textbf{31.35}/\textbf{30.28}} & \revise{\textbf{30.83}/\textbf{29.81}} \\
\hline
\end{tabular}}
\vspace{-0.1in}
\end{table}

\subsection{Scalability Comparison}
To evaluate the scalability of all methods, we train different models by varying the training data size. In particular, we sample 20\%, 40\%, 60\%, 80\% and 100\% from the training data of \textit{Chengdu} and \textit{Xi'an}, and collect the associated MAPE of the online prediction over the test data, to more intuitively reflect the performance of the different methods. From Table~\ref{tab:sca}, we have the following observations:

\noindent (1) As the volume of data increases, the performance of all methods improves, indicating that deep learning models have not yet reached their learning capacity limits. Larger datasets provide more diverse spatio-temporal information, which helps models better capture and understand the patterns and features in the data, thereby enhancing the models' ability to generalize effectively.

\noindent (2) Deep learning methods are more stable and effective compared to \textbf{Avg}. For instance, the MAPE of \textbf{Avg} on the \textit{Chengdu} dataset increases by $\frac{58.78 - 47.35}{47.35} = 24.14\%$ when we only use 20\% training data, while the ratio is only $\frac{33.97 - 30.83}{30.83} = 10.18\%$ for \textbf{\modelName}.

\noindent (3) Compared to all other baseline methods, our approach, \textbf{\modelName}, consistently achieves the best performance across all scales of data.

\begin{figure}
  \centering  
  \includegraphics[width=0.85\linewidth]{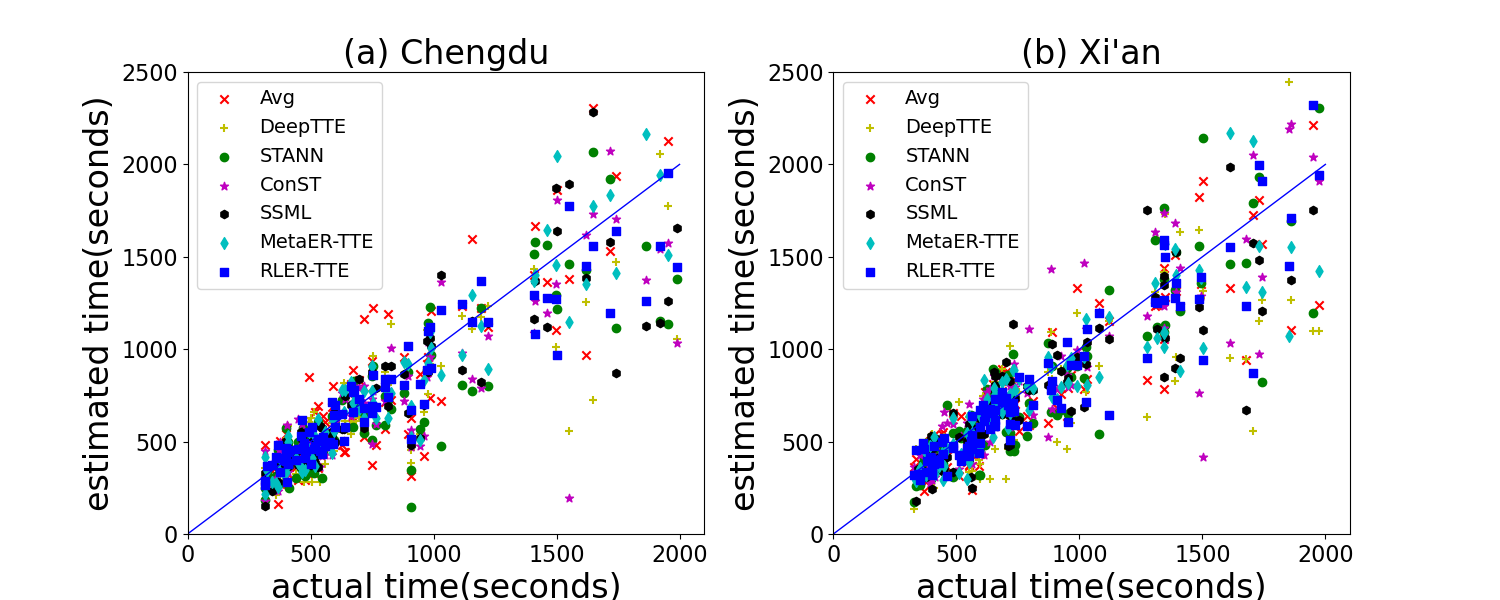}
  \vspace{-0.15in}
  \caption{Estimated time vs. actual time.}
  \label{fig:case}
  \vspace{-0.15in}
\end{figure}

\begin{figure}
  \centering  
  \includegraphics[width=0.85\linewidth]{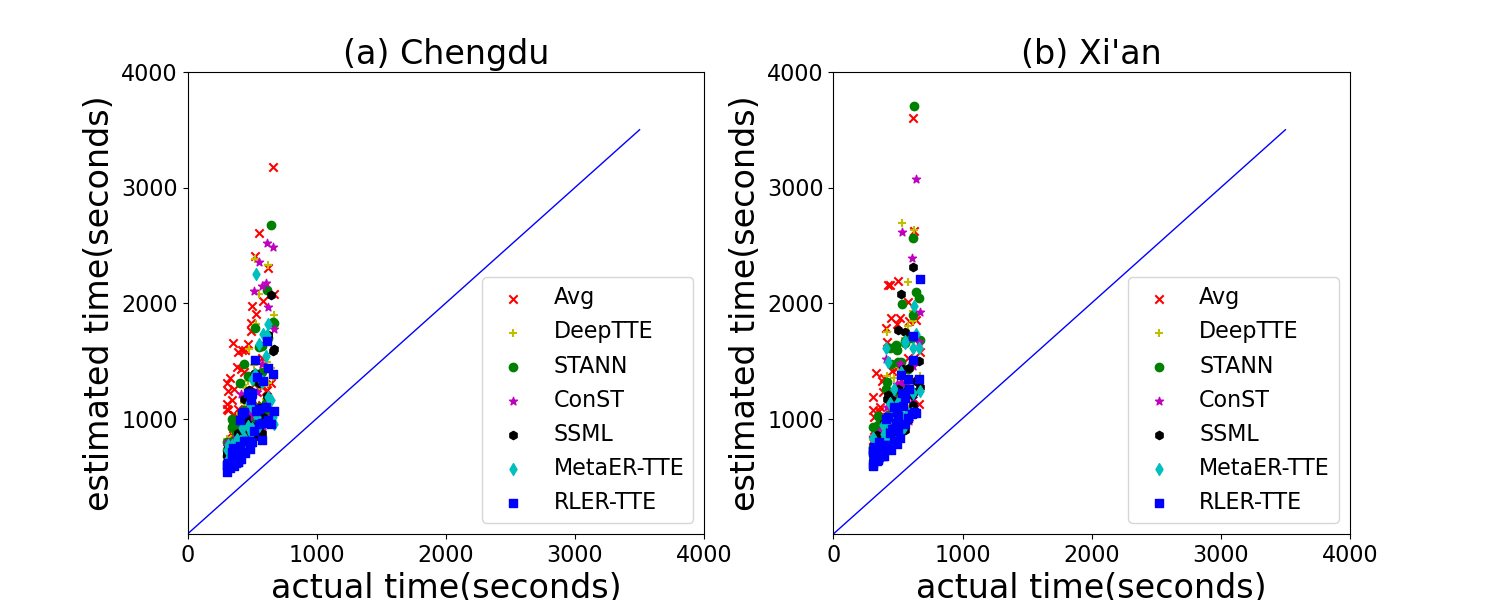}
  \vspace{-0.15in}
  \caption{Estimated time vs. actual time on MAPE.}
  \vspace{-0.1in}
  \label{fig:case_mape}
\end{figure}

\subsection{Case Study}
To visually demonstrate the effectiveness of our framework, we employ scatter plots to compare all methods, with the \( y \)-axis representing the estimated time and the \( x \)-axis representing the actual time. As illustrated in Figure~\ref{fig:case}, we include a reference line \( y = x \) for comparison. The closer the data points are to this reference line, the more accurate the predictions are. Conversely, the further the data points deviate from this line, the worse the performance of the method. We randomly sample 100 test data points from the \textit{Chengdu} and \textit{Xi'an} datasets. Our observations are
as follows: 

\noindent (1) The average error of the \textbf{Avg} method is generally larger but does not produce outliers with significantly high errors. In contrast, deep learning methods occasionally exhibit anomalous cases with substantial errors. (2) Across data of all lengths, our method, \textbf{\modelName}, demonstrates closer adherence to the reference line compared to other methods. (3) As actual time increases, the errors for all methods grow, but \textbf{\modelName} has the smallest degree of increase.

To evaluate the performance of each method under worst-case scenarios, we further select the 50 worst-performing instances for each method based on MAPE scores. These instances are plotted in a similarly formatted scatter plot in Figure~\ref{fig:case_mape}. Since our selection is based on MAPE scores, shorter actual times result in larger relative errors, causing the majority of cases to cluster on the left side. Similarly, the performance of each method can be judged by the proximity of the data points to the reference line. We can find that: 

\noindent (1) It is visually apparent that \textbf{\modelName} adheres more closely to the reference line. (2) The \textbf{Avg} method performs worse in the worst-case scenario, with its errors being significantly larger than those of deep learning methods.


\vspace{-0.05in}
\section{RELATED WORK}

\subsection{Travel Time Estimation}
Travel Time Estimation (TTE) plays a crucial role in intelligent transportation systems, serving as the foundation for tasks such as traffic forecasting and navigation planning. Current research in this field can be broadly categorized into two types: offline travel time estimation and online travel time estimation.

\noindent \textbf{Offline travel time estimation.} Offline travel time estimation primarily involves learning and evaluation on datasets without considering real-time prediction scenarios. The mainstream studies
 can be primarily divided into two categories: \textit{OD-based} and \textit{path-based}. The \textit{OD-based} methods~\cite{STNN, TOPK, TITS19, RTODT, MURAT} take only the origin, destination locations, and departure times as inputs, and estimate the travel time for the corresponding OD pairs. Yuan et al.~\cite{DeepOD} consider historical trajectory data associated with OD pairs and propose an efficient encoding technique based on neural networks, while Lin et al.~\cite{ODT} utilize diffusion models to further explore the complex relationships between trajectory data and OD pairs. However, these methods fail to capture the rich spatio-temporal information within the travel route, making it difficult to perform tasks such as navigation planning that require travel time estimation for a specified route. Therefore, the \textit{path-based} methods~\cite{kdd14, DeepTravel, vldb20, ConST, MetaTTE, RTTE}, which take a travel route as input, are proposed to address this issue. Luan et al.~\cite{vldb20} consider future traffic information and propose a deep learning system that integrates both a future traffic flow prediction model and a travel time estimation model~\cite{DeepTTE}. Zhao et al.~\cite{HRA} focus on aggregation operations in travel time estimation and design an Inverted Table to record aggregation requests. They propose a hierarchical reduction architecture suitable for multiple deep learning models. Yuan et al.~\cite{RTTE} focus on the unique correlations within link data, emphasizing characteristics such as heterogeneity and proximity between links. They treat traffic conditions as dynamic data, processing them separately from static link data. However, the so-called dynamic data is actually simulated based on different historical traffic conditions, making it essentially an offline attribute of the road network. Offline TTE studies typically perform prediction once before the journey begins, considering only the internal characteristics of static data in offline scenarios and primarily focusing on technical accuracy, while neglecting the real-time data collected during the journey in online scenarios. From a data perspective, our work emphasizes the characteristics of real-time data and the correlation between real-time and static data. Furthermore, we identify that real-time data exhibits characteristics such as unhelpful, misleading, and meaningless. Based on these findings, we innovatively propose an efficient online decision-making framework that redefines the data flow of the ER-TTE task and scientifically designs a Markov Decision Process to address this challenge.



\noindent \textbf{Online Travel time estimation.} Online travel time estimation is oriented towards online scenarios and is specifically designed to address the unique requirements of real-time environments. Kristin et al.~\cite{latte} proposed a stream-archive query system, which combines real-time traffic data streams and large transportation data archives for real-time travel time estimation. Lint et al.~\cite{OLTTE} utilize simple neural networks to estimate travel time and employ the Extended Kalman Filter (EKF) to update the model weights online, enabling adaptation to changes in real-time traffic data. However, these methods are more reliant on the quality of historical data and lack the powerful capabilities of deep learning models. Therefore, Fu et al.~\cite{compactETA} propose an online prediction framework consisting of a real-time prediction MLP and a sophisticated offline asynchronous updater. Han et al.~\cite{iETA} introduce an incremental learning framework designed to continuously explore the spatio-temporal correlations present in new data. However, these methods only concentrate on the routes themselves and are merely extensions of offline methods applied in online scenarios, neglecting the user behavior patterns collected in real-time during the driving process. To address this problem, recent work has focused on en route travel time estimation. Fang et al.~\cite{SSML} first propose the ER-TTE task and formulate it as a few-shot learning problem, employing a model-based meta-learning approach to predict travel time. Fan et al.~\cite{MetaER-TTE} improved upon meta-learning techniques by personalizing the initialization based on the unique context of each trajectory. However, existing methods fail to consider the complexity of real-world scenarios and do not sufficiently explore the relationship between driving behavior data and route data. Moreover, in the experimental settings, they only simulate a single request with a fixed traveled route, lacking a reasonable training and evaluation design. To address these issues, our work redefines the data pipeline and proposes a novel framework comprising a \textit{Decision Maker} and a \textit{Predictor}. Furthermore, we design comprehensive experiments and evaluations, establishing a new standard for the ER-TTE task.

\vspace{-0.05in}

\subsection{Deep Learning in Spatio-Temporal Data}

Recently, the remarkable performance of deep learning techniques in the field of spatio-temporal data management and mining has garnered attention. First, attention-based methods have shown advanced performance in capturing complex spatio-temporal correlations, with applications in areas such as Next POI Recommendation~\cite{STAN, JNPOI} and traffic flow prediction~\cite{DeepTP, attTFP, DBLP:conf/cikm/Chen0JBW24,DBLP:journals/dase/YuanL21}. Second, an increasing number of methods employ Graph Neural Networks (GNNs) and hybrid architectures combining various neural networks to analyze spatio-temporal data. For example, Yuan et al.~\cite{Nuhuo} utilized GNNs and RNNs to capture spatio-temporal correlations for Traffic Speed Histogram Imputation, while Wu et al.~\cite{TSF} used Graph Convolution Networks (GCNs) and Graph Isomorphism Networks (GINs) to extract temporal dependencies within time series and spatial correlations between different time series for effective time series prediction. Last but not least, numerous deep learning approaches based on other architectures have also been applied to spatio-temporal data processing. For instance, Ahuja et al.~\cite{STDP} employ Variational Auto Encoders (VAE) to handle the release and denoising of spatio-temporal data, capturing and reconstructing complex spatio-temporal patterns.

\vspace{-0.06in}

\section{CONCLUSION}
This paper has introduced \textbf{\modelName}, a novel and efficient framework designed to address the complex challenge of en route travel time estimation. Firstly, we redefined the implementation path of ER-TTE, proposing for the first time a novel pipeline comprising a \textit{Decision Maker} and a \textit{Predictor}. Next, we meticulously designed a Markov Decision Process and deployed efficient reinforcement learning to achieve effective autonomous decision-making in dynamic online environments. Furthermore, we established the training and evaluation standards for ER-TTE and introduced a curriculum learning strategy to manage the large volume of training data. Extensive experiments on real-world datasets validated the effectiveness, efficiency, and robustness of our model.

\begin{acks}
This work was supported by the National Natural Science Foundation of China (No.62032003, No.62425203) and was also supported by Beijing Natural Science Foundation (No.L212032). Haitao Yuan is the corresponding author of the work.
\end{acks}

\bibliographystyle{ACM-Reference-Format}
\bibliography{ref}










\end{document}